\documentclass[10pt,twocolumn,letterpaper]{article}

\PassOptionsToPackage{table,dvipsnames}{xcolor}
\usepackage{wacv}              

\usepackage{booktabs}     
\usepackage{amssymb}      
\usepackage{pifont}       

\newcommand{\cmark}{\ding{51}}  
\newcommand{\xmark}{\ding{55}}  

\definecolor{wacvblue}{rgb}{0.21,0.49,0.74}
\usepackage[pagebackref,breaklinks,colorlinks,allcolors=wacvblue]{hyperref}

\def\wacvPaperID{1069} 
\def\confName{WACV}
\def\confYear{2026}

\def\NAME{SAVeD} 

\makeatletter
\renewcommand*{\@fnsymbol}[1]{%
  \ifcase#1 \or =\or *\or \dagger\or \ddagger\or \mathsection\or
  \mathparagraph\or \|\or **\or \dagger\dagger\or \ddagger\ddagger
  \else\@ctrerr\fi}
\makeatother

\makeatletter
\renewcommand{\and}{\unskip\hspace{3em plus .5em minus .5em}}
\makeatother

\title{SAVeD: Learning to Denoise Low-SNR Video for Improved Downstream Performance}

\author{
  Suzanne Stathatos \and
  Michael Hobley \and
  Pietro~Perona\thanks{Equal advising contribution} \and
  Markus~Marks$^{=}$\\
  Caltech\\
  {\tt\small \{sstathat, mahobley, perona, marks\}@caltech.edu}
}

\begin{document}
\maketitle
\setcounter{footnote}{0}
\begin{abstract}
    Low signal-to-noise ratio (SNR) videos—such as those from underwater sonar, ultrasound, and microscopy—pose significant challenges for computer vision models, particularly when paired clean imagery for denoising is unavailable.
    We present Spatiotemporal Augmentations and denoising in Video for Downstream Tasks (\NAME{}), a novel self-supervised method that denoises low-SNR sensor videos using only raw noisy data. By leveraging distinctions between foreground and background motion and exaggerating objects with stronger motion signal, \NAME{} enhances foreground object visibility and reduces background and camera noise without requiring clean video. \NAME{} has a set of architectural optimizations that lead to faster throughput, training, and inference than existing deep learning methods. We also introduce a new denoising metric, FBD, which indicates foreground-background divergence for detection datasets without requiring clean imagery. Our approach achieves state-of-the-art results for classification, detection, tracking, and counting tasks and it does so with fewer training resource requirements than existing deep-learning-based denoising methods. Project page \href{https://suzanne-stathatos.github.io/SAVeD}{here}, Code: \href{https://github.com/suzanne-stathatos/SAVeD}{https://github.com/suzanne-stathatos/SAVeD}.
\end{abstract}

\section{Introduction}
\label{sec:intro}
Motion may be the only way to identify objects in video with low signal-to-noise-ratio (SNR), camouflage, or complex textures that may hinder frame-by-frame object detection. The human visual system is excellent at capturing observable motion~\cite{giese2003neuro}, a capability which has not yet been reproduced by modern computer vision models. Learning to exploit motion cues will improve models' abilities to detect and track objects of interest in noisy video.

In several scientific \cite{Holste2024} and medical applications \cite{sundararajan2018DL4Bio, dargan2020DL4Applications}, clean (noise-free) imagery are not available to train image or video denoisers. One attractive element of self-supervised models is their ability to find the signal within the noisy imagery itself. Self-supervision can, too, be more robust and generalizable to various noise type \cite{shi_how_2022, marks2024closer}.

This work aims to enhance motion signals in low-signal-to-noise (SNR) data with non-stationary backgrounds, such as underwater, ultrasound, and sonar videos, to improve downstream supervised classification, detection, and tracking. We address sensor, background, and noise challenges with \NAME{}, a self-supervised learning method to boost signal in noisy video. We exploit object motion to boost the SNR across frames. Inspired by work on self-supervised reconstruction \cite{khalil2024learning, sun2022bkind}, self-supervised video understanding \cite{recasens2021broaden}, and anomaly detection  \cite{yang2023anom, liu2018anom}, we use an encoder to encode appearance frames, a temporal bottleneck, and a decoder network to reconstruct the denoised frame. 

\begin{figure}[tp]
    \centering
    \includegraphics[width=\linewidth]{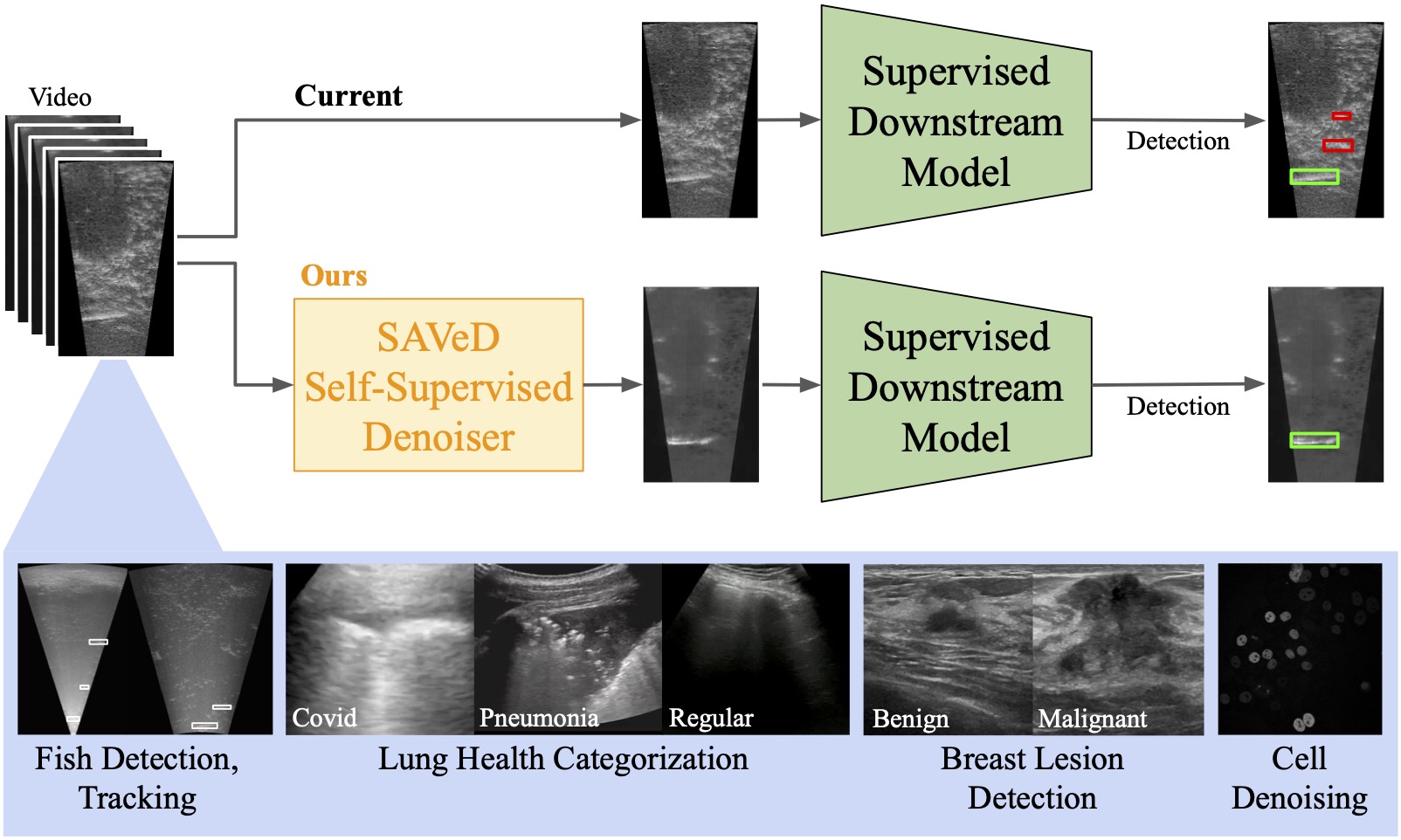}
    \caption{\textbf{SpatioTemporal Denoising improves classification, detection, tracking, and counting in video.} 
    We denoise sonar and ultrasound videos of fish in a river, lung scans, breast lesion scans, and cell microscopy to improve downstream classification, detection, tracking, and counting tasks. 
    We propose a self-supervised method to enhance the foreground signal of video frames without manual annotations or clean imagery. 
    Our method works on grayscale videos with: non-stationary backgrounds, low signal-to-noise-ratios, and a variable number of objects in a video.}
    \label{fig:fig1}
\end{figure}


\noindent Our main contributions are:
\begin{itemize}
    \item We propose \NAME{}, a novel, state-of-the-art, resource-efficient self-supervised method to improve the signal in low-SNR video with variable numbers of agents and non-stationary background; details are in \cref{sec:method}. 
    \item We introduce a novel denoising metric, $FBD$, that does not rely on clean imagery; it instead relies on background-foreground appearance distribution differences, in \cref{sec:method}.
    \item We propose a rich benchmark for low-SNR video denoising consisting of a diverse collection of low-SNR video domains (sonar video of fish, ultrasound video of lungs and breast tissue, and microscopy video) and a diverse collection of downstream visual tasks (classification, detection, tracking, and counting), in \cref{sec:experiments}.
\end{itemize}
\begin{figure*}[!htp]
    \centering
    \includegraphics[width=0.8\linewidth]
    {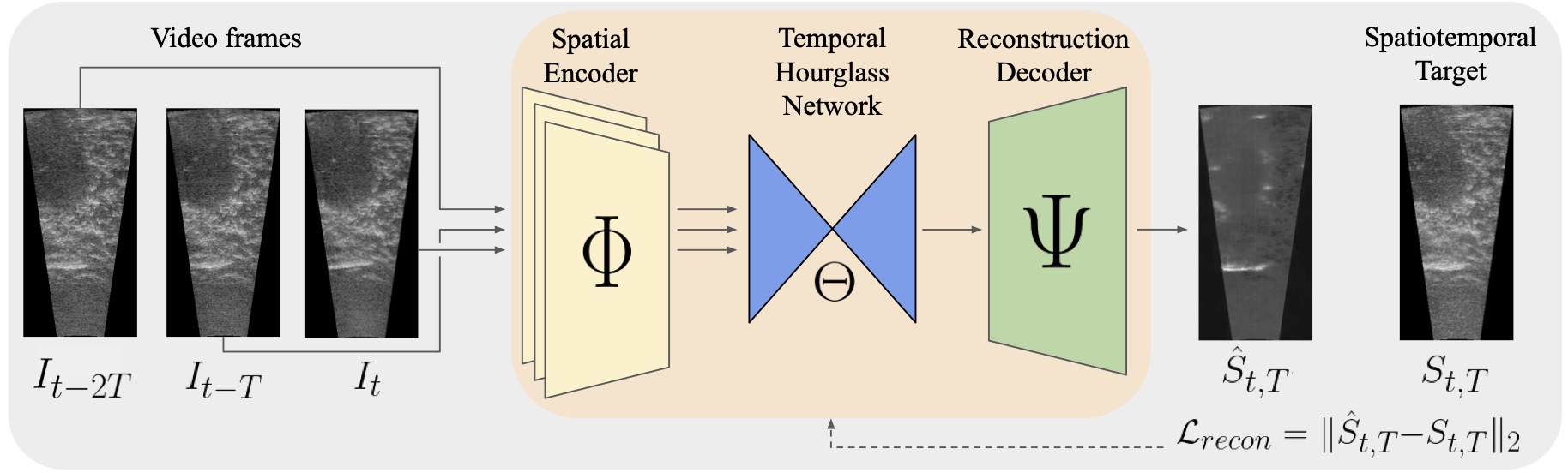}  
    \caption{\textbf{\NAME{}, our approach for self-supervised denoising using spatiotemporal difference and identity reconstruction}. $I_t$, $I_{t-T}$, and $I_{t-2T}$ are video frames at times t (current frame), t-T, and t-2T. These frames are input to an appearance encoder $\Phi$. The resulting feature representations are input to a spatiotemporal bottleneck $\Theta$ that compresses the 3 appearance features into a single spatiotemporal feature representation. Our model then predicts the reconstruction target, defined in \cref{eqn:reconstruction_target} in \cref{sec:reconstruction}, using the reconstruction decoder $\Psi$. The loss, defined in \cref{eqn:recon_loss}, is calculated and backpropagated through all networks. The architecture is discussed in more detail in \cref{sec:method_arch}.}
    \label{fig:fig2}
\end{figure*}

\section{Related Work}
\label{sec:related_work}

Several works in the image and video denoising space require paired clean/noisy imagery for training, evaluation, or both, limiting their applicability in real-world denoising scenarios \cite{zhang2017beyond, li2023ntire, jiang2025efficient, marcosmorales2023evaluating}. We, notably, do not, and we focus this section on existing self-supervised and unsupervised approaches.

\noindent{}\textbf{Self-supervised and unsupervised image (frame-by-frame) denoising.} Several approaches use variants of blind-spot networks or pixel-wise masking to denoise imagery, including frame-by-frame video. Noise2Self \cite{batson2019noise2self} and Noise2Void (N2V) \cite{krull2019noise2void} train on noisy images without requiring clean targets or paired data. N2V trains a blind-spot network to predict masked pixels' intensities from neighboring pixels. Others \cite{laine2019high, batson2019noise2self, jang2023selfsupervised} refrain from masking pixels via structural blind-spot networks with half-plane receptive-field U-Nets \cite{ronneberger2015unet}.  \citet{jang2023selfsupervised} use a conditional blind-spot network and a loss that regularizes denoised images without masking input pixels. Neighbor2Neighbor \cite{huang2021neighbor2neighbor} proposes a self-supervised loss between two sub-sampled images. In general, noise in real-world imagery, including acoustic imagery, has unknown or non-stationary statistics that are spatially correlated, violating assumptions of pixel-wise independence. 

\noindent{}\textbf{Video Denoising with Flow Estimation}
Some video denoising methods leverage videos' spatiotemporal structure by using optical flow for motion compensation \cite{xue2019video, tassano2020dvdnet}. DVDnet \cite{tassano2020dvdnet} uses calculated flow-estimates to manually warp frames, align their contents, and process them collectively with a CNN. VDFlow \cite{wang2020vdflow} jointly learns video deblurring and optical flow. These approaches target video where one component is moving (\eg an object or the camera). However, when objects' motion affects the background's motion, or when objects are small, we find that leveraging optical flow enlarges objects' motion and makes it difficult to distinguish multiple closely-located objects. An example is in \cref{fig:RAFT_on_cfc} in the Supp Mat.

\noindent{}\textbf{Video Restoration and Denoising with Spatial and Temporal Consistency}
UDVD \cite{Sheth_2021_ICCV} uses a patch-wise noise-to-noise training strategy to predict clean frames by estimating masked pixels from adjacent neighborhoods of noisy frames. It relies on temporal redundancy across frames and does \textit{not} rely on clean imagery, making it appealing for in-the-wild data. However, UDVD assumes gaussian noise, while underwaters sonar and ultrasound videos have pink and speckle noise  \cite{zhang2021seasonal_ambient_noise, zhou2023sonar_speckle, despeckling2013}. UDVD's patchwise nature may cause the model to overwhelmingly learn background when most of the video is spatially and temporally dominated by background. UMVD \cite{UMVD} extends UDVD to focus on microscopy data. It uses frame-level augmentations and a reconstruction loss that predicts each frame from its noisy temporal neighbors. It assumes that the signal is consistent across frames while the noise is not, and it assumes smooth object motion. However, like UDVD, it is also patch-wise. 

AverNet \cite{zhao2024avernet}, another self-supervised video restoration model, similarly relies on temporal consistency to fix time-varying unknown degradations. It uses two modules: 1.) prompt-guided alignment to line up video frames at the pixel level and 2.) prompt-conditioned enhancement, which restores each frame by adapting to the image's degredation. The method was tested and performed well on videos with varying levels of gaussian, poisson, and speckle noise, gaussian and resizing blur, and jpeg and video compression. 

LG-BPN \cite{wang2023lgbpn}, another patch-wise denoising method, uses two main ideas: local guidance and internal consistency. Local guidance preserves image structures like edges or textures that are consistent even with noise. Internal consistency assumes nearby patches in space and time share similar structure. However, LG-BPN struggles with long-range temporal dependency and highly-structured noise, such as sensor artifacts. RVRT \cite{liang2022rvrt} is a transformer-based model for denoising, deblurring, and super-resolution. It processes videos frame-by-frame, but models long-term dependencies (remembering what it learned from frames that are further apart) using spatiotemporal attention. It, however, requires very large memory even for inference.

\noindent{}\textbf{Denoising autoencoders (DAEs)} were originally introduced to learn robust representations. During training, DAEs intentionally add noise to input data and learn to reconstruct the original uncorrupted signal. mDAE \cite{mDAE}, a method for missing data imputation (replacing missing or unavailable data), improves performance on several datasets. 
DAEs have also been applied to video tasks. CompDAE \cite{compDAE} explicitly models noise from snapshot compressive imaging measurements in low-light conditions to improve edge detection and depth estimation. TADA \cite{Choi2025TADA} uses an adversarial denoising autoencoder to remove EMG noise from EEG time series data. Our work similarly extends the application of DAEs to spatio-temporal grayscale video denoising; we uniquely combine temporal frames to enhance signal quality while simultaneously addressing the increased noise introduced by this process.


\textbf{Sonar and Ultrasound}
Sonar and ultrasound present challenges to the computer vision community. Key characteristics include pink noise, brighter pixel intensities for non-cavity objects compared to the background, and limited distinctiveness from appearance features. \citet{weld2025standardisationconvexultrasounddata} address ultrasound sensor's variability via geometric analysis and augmentation. Unlike many computer vision datasets based on signal from light intensity, ultrasound, sonar,  lidar, and radar rely on wave echoes. The ``camera'' emits waves that reflect off objects and return to the sensor; distance is measured by the echo's return time. Sonar and ultrasound use sound waves (mainly in liquids), while lidar uses laser pulses and is common in air and land environments \cite{Geiger2013VisionMR, caesar2020nuscenes, chang2019argoverse}. We focus on one sonar dataset and two ultrasound datasets described in more detail in \cref{sec:datasets}. 

\section{Method}\label{sec:method}

The goal of \NAME{} (\cref{fig:fig2}) is to enhance objects' signal by isolating and emphasizing their motion in video with a non-stationary, fluid background. Inspired by prior work \cite{sun2022bkind, jakab2018conditionalImageGen, ryou2021weaksupKPdisco}, we use an encoder-decoder framework. Our contribution is twofold: (1) a reconstruction target based on spatiotemporal differences across neighboring frames, and (2) a denoising metric that does not require clean ground truth. The method assumes that background spatiotemporal statistics differ from those of the foreground. 

\subsection{Self-supervised denoising}
In our benchmarks' low-SNR videos, signals are distributed across time; as such, we want to condense information from multiple timesteps into a single frame to exaggerate the signal. We do this through the reconstruction target. For simplicity, we choose to reconstruct the spatiotemporal combination of 3 frames, expressed, in \cref{eqn:reconstruction_target}, from three input frames, $I_t$, $I_{t-T}$, and $I_{t-2T}$. We explore a vanilla autoencoder, UNets with and without skip connections and residual layers, and 3D convolutions (in \cref{tab:big_table}), and find a simple encoder-bottleneck-decoder framework optimal. We also explore $N>3$ input frames, $T>1$, the target without the DAE network as input to downstream tasks, and more in our ablations in the Supplemental \cref{tab:ablation_reconstruction_targets}, \ref{tab:additional_ablations}, and \cref{fig:reconstruction_targets_supp}. 

We use an encoder-decoder architecture, seen in \cref{fig:fig2}, with a spatial encoder, $\Phi$, a temporal hourglass network, $\Theta$, and a reconstruction decoder, $\Psi$. During training, spatial encoders, $\Phi$, take $I_t$, $I_{t-T}$, and $I_{t-2T}$ as input to generate spatial feature embeddings, which are then used by the hourglass network, $\Theta$, to generate a spatiotemporal feature embedding; this embedding passes to $\Psi$ to reconstruct the learning objective $\hat{S}_{t,T}$.
\begin{equation}
\hat{S}_{t,T}=\Psi(\Theta(\text{concat}(\Phi(I_t), \Phi(I_{t-T}), \Phi(I_{t-2T}))))
\end{equation}

\subsection{Reconstruction target and loss}\label{sec:reconstruction}

\textbf{Target.} 
We use the \textit{directionally-positive frame difference with the current frame (PFDwTN)} as our reconstruction target. This combines the current frame with the positive motion from the previous and next frames. Directionally-positive motion of the next frame is defined: $\max(0, I_{t+T} - I_t)$; of the previous frame, it is defined: $\max(0, I_{t-T} - I_t)$. In both cases, motion is relative to the current frame $I_t$. Note that the previous frame $I_{t-T}$ goes into the network, while the future frame $I_{t+T}$ does \textit{not}. It is used only when calculating the ground-truth target for the loss.
 
\looseness-1 
To handle frames where the background movement does not differ significantly from the foreground objects' motion (\ie, stationary objects), we include the original frame, $I_t$, in the reconstruction target. The overall target is:
\begin{equation}\label{eqn:reconstruction_target}
S_{t,T} = \max(0, I_{t-T} - I_t) + I_t + \max(0, I_{t+T} - I_t)
\end{equation}

Other motion-augmenting targets that we tested are defined/visualized in Sec. \ref{sec:ablation_additions} and \cref{fig:reconstruction_targets_supp} in the Supp. Mat.

\noindent{}\textbf{Loss}. We apply mean-squared-error loss for reconstructing the current frame with augmented motion signatures:
\begin{equation}\label{eqn:recon_loss}
\mathcal{L}_{recon} = \|\hat{S}_{t,T}-S_{t,T}\|_2
\end{equation}

\subsection{Noise Removal Network}\label{sec:method_arch}

\noindent{}\textbf{Appearance Encoder $\Phi$}. We implement a 6-layer CNN. Each layer consists of a convolutional block (Conv2D + ReLU) followed by max pooling, progressively increasing the number of feature channels while reducing the spatial dimension, $\mathbb R^{(H,W,1)} \rightarrow \mathbb R^{(\frac{H}{32},\frac{W}{32}, 512)}$. We also save skip connections (sequential max pools and 1x1 convolutions) to be used by the hourglass network and decoder. This design uses a fraction of the parameters and FLOPs from an off-the-shelf UNet, to let the network capture multi-scale features efficiently. See \cref{tab:denoiser_time} for efficiency comparisons.

\noindent{}\textbf{Temporal Hourglass Network $\Theta$} is an hourglass network with a bottleneck consisting of two 3x3 convolutional layers with 512 channels, each followed by ReLU activation. We also have skip connections as feature combiners at each level of the network, designed to merge information from the provided appearance features' skip connections.

\noindent{}\textbf{Reconstruction Decoder $\psi$} has 6 upsampling stages, each consisting of a ConvTranspose followed by convolutions and ReLU activations. At each layer, the skip connections from the corresponding encoder level are concatenated with the upsampled features. The decoder reduces the number of channels while increasing the spatial dimension ending with a single-channel output, $\mathbb R^{(\frac{H}{32},\frac{W}{32}, 512)} \rightarrow \mathbb R^{(H,W,1)}$.

More details on each of these modules is in the Supplemental materials \cref{tab:denoiser_arch}.


\subsection{Denoising Metric}\label{sec:denoising_metric}
Typically \cite{krull2019noise2void, laine2019high, jang2023selfsupervised, huang2021neighbor2neighbor, Sheth_2021_ICCV}, denoising networks use Peak Signal-to-Noise Ratio (PSNR) and Structural Similarity Index Measure (SSIM) \cite{wang2004ssim} as evaluation metrics. PSNR is the ratio of maximum signal power to noise power, and SSIM measures perceived image quality. Both metrics rely on having clean imagery for comparison. Our denoising approach is unsupervised -- we do \textit{not} have clean imagery. As a result, we design a new metric, Foreground-to-Background Divergence ($FBD$), to evaluate denoised performance when we have detection annotations, and we rely on downstream performance as a proxy for denoised performance on datasets with different task annotations.

\subsubsection{Foreground-to-Background Divergence (\textit{FBD}) Metric for Unsupervised Denoising} \label{sec:KL} 
Recall that we assume that our \textit{downstream} models are supervised. Therefore, for detection tasks, we can assume we have bounding boxes or segmentation masks. For simplicity, we call detection annotations ``boxes'', though the same approach works for segmentation masks. 

\begin{figure}
    \centering
    \includegraphics[width=0.5\textwidth]{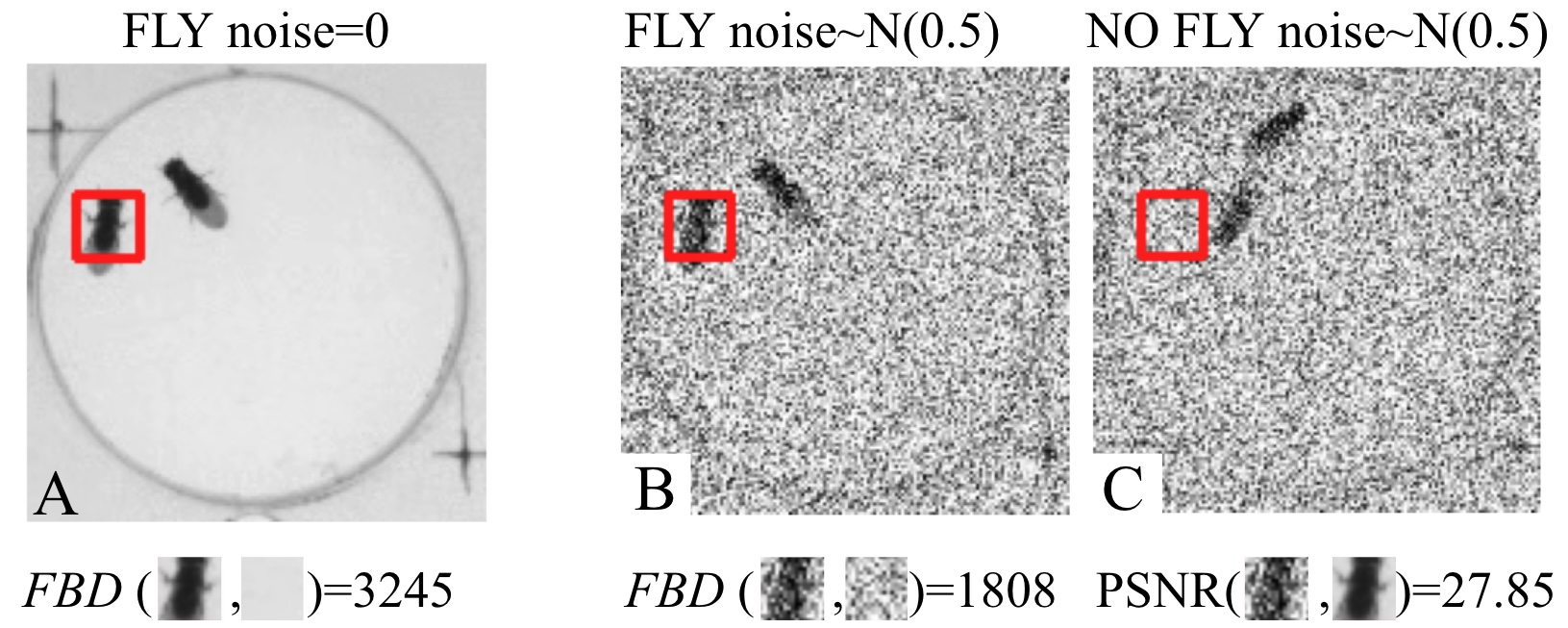}
    \caption{High-SNR visualization and calculations for $FBD$ and PSNR on Fly-vs-Fly \cite{flyvfly} with synthetic noise. In order to calculate the PSNR of \textbf{(B)}, a `clean' image \textbf{(A)} is needed. \textbf{(A)} is not needed for $FBD$, though \textbf{(C)} is.\vspace{-2mm}}
    \label{fig:fbd_example}
\end{figure}

$FBD$ measures how well a denoising method makes objects distinguishable from background by computing the KL-Divergence between a region containing an object (foreground) and the same region at a different time without it (background). An example is shown in \cref{fig:fbd_example}. Unlike PSNR, which requires a clean reference, $FBD$ does not, which is critical in many real-world applications.

For each object's box $b$ in each frame $I$ of each video, we take the density of pixel intensity values: $d_{b} = I[b]$, where $b$ are the indices associated with the box. Then, we take the density of pixel intensity values from a different frame $\tilde{I}$ of the \textit{same} video at the same box location $b$ where we know there is no object, $\tilde{d}_b$. If the distributions are separable \ie, the distribution of object pixels is distinct from the distribution of background pixels, then the denoising method works as intended. We measure the separability between object and non-object via the Kullback-Leibler (KL) Divergence \cite{Kullback51klDivergence}. KL divergence measures the distance between two probability distributions P and Q as follows:
\begin{equation}
D_{KL}(P\|Q) = \int p(x) \log(\frac{p(x)}{q(x)}) dx
\end{equation}
To generate a metric for a data split, we average the $D_{KL}$ over all N bounding boxes to get 
\begin{equation}\label{eqn:fbd}
FBD_{D_{KL}} = \frac{1}{N}\sum_{b\in{N}} D_{KL}( d_b \|\tilde{d}_b)
\end{equation}
A visualization of this metric can be seen in Supplementary materials  Fig.\ \ref{fig:denoiser_metric_details}. A larger score (divergence) indicates greater distinguishability of objects from background.
\section{Experiments}\label{sec:experiments}
We demonstrate that \NAME{} can improve performance in low-SNR videos across medical and ecological applications (\cref{sec:results}). We evaluate our processed images on downstream tasks for detection, tracking, counting, and classification. 

\subsection{Datasets} \label{sec:datasets}

\textbf{Caltech Fish Counting 2022}  (CFC22) \cite{cfc2022eccv} is designed for detection, tracking, and counting fish in low-signal-to-noise sonar video. This dataset contains 1,567 sonar videos from seven different cameras on three rivers in Alaska and Washington. The videos are grayscale, their resolutions range from 288x624 to 1,086x2,125, their frame rates range from 6.7 to 13.3 fps, and each video is on average 336 frames (38s) in duration \cite{cfc2022eccv}. In total, there are 527,215 frames with 8,254 unique fish, totaling 516k bounding boxes and 16.7 hours of video \cite{cfc2022eccv}.  The dataset includes significant domain shifts (\eg, background topology, occlusion, fish densities, fish sizes, camera noise), requiring models to generalize effectively across conditions. 

\noindent{}\textbf{Breast Lesion Ultrasound Video Dataset} \cite{CVANet} (BUV) is designed for classifying (benign or malignant) and localizing breast lesions. The dataset contains 188 videos, of which 113 are malignant and 75 are benign. These videos collectively have 25,272 images, each with 1 detection; the number of ultrasound images in each video range from 28 to 413. Each video has a complete scan of the abnormal tissue. The dataset has a random train--test split of 150--38 videos respectively\cite{CVANet}.

\noindent{}\textbf{The Point-of-care Ultrasound dataset (POCUS)} \cite{born2021accelerating, born2021l2} contains convex and linear probe lung ultrasound images and videos for classifying COVID-19 and pneumonia. It includes 247 videos and 59 images; we use only the videos. Of these, 70 show COVID cases, 45 \textit{possible} COVID, 51 bacterial pneumonia, 6 viral pneumonia, and 75 healthy lungs. Videos are sampled at 10Hz, and frames are grouped by video as in Born et al. \cite{born2021accelerating, born2021l2}. In total, we extract 9,184 frames with an average size of 499×463 pixels.

\noindent{}\textbf{Fluorescence microscopy dataset} \cite{ulman2017objective} (Fluo) is a dataset of fluorescence-microscopy recordings of live cells in \cite{ulman2017objective}. We use the same videos as UDVD \cite{Sheth_2021_ICCV}: Fluo-32DL-MSC (CTC-MSC), of mesenchymal stem cells, and Fluo-N2DH-GOWT1 (CTC-N2DH), of GOWT1 cells. This dataset also contains no ground-truth clean data. There are a total of 560 frames and four videos. 

\subsection{Training Procedure -- Denoiser}
We train \NAME{} using the reconstruction objective in \cref{eqn:recon_loss}. During training, we rescale CFC22 and POCUS images to 1024x512 and BUV and Fluo images to 1024x1024. For POCUS, we use a sample rate of 10Hz, as that is what the downstream process uses. For all other datasets, we use all frames. Because SAVeD is self-supervised, when training the \textit{denoiser} for each dataset, we train over all data. We train for 20 epochs for CFC22, 120 epochs for POCUS, 40 epochs for BUV, and 1000 epochs for Fluo; we found these numbers of epochs sufficient for training to converge. These took 20 hours, 0.5 hours, 2 hours, and 2 hours, respectively, on 2 RTX 4090 GPUs; this is less time than other network-based denoising methods as seen in \cref{tab:denoiser_time}. Additional details, including hyperparameter configurations, are in the Supplemental Materials \cref{sec:our_denoiser_hyperparams_and_details}.

After training \NAME{}, we generate denoised frames for all splits. In the case of CFC22, we combine the denoised image as two channels and the background-subtracted frame, $(I_v)_t - \bar{I_v}$, as the last channel. For POCUS, BUV, and Fluo, we combine the denoised image as two channels and the median-filtered image as the last channel. 
\begin{figure*}[t]
    \centering
    \includegraphics[width=\linewidth]{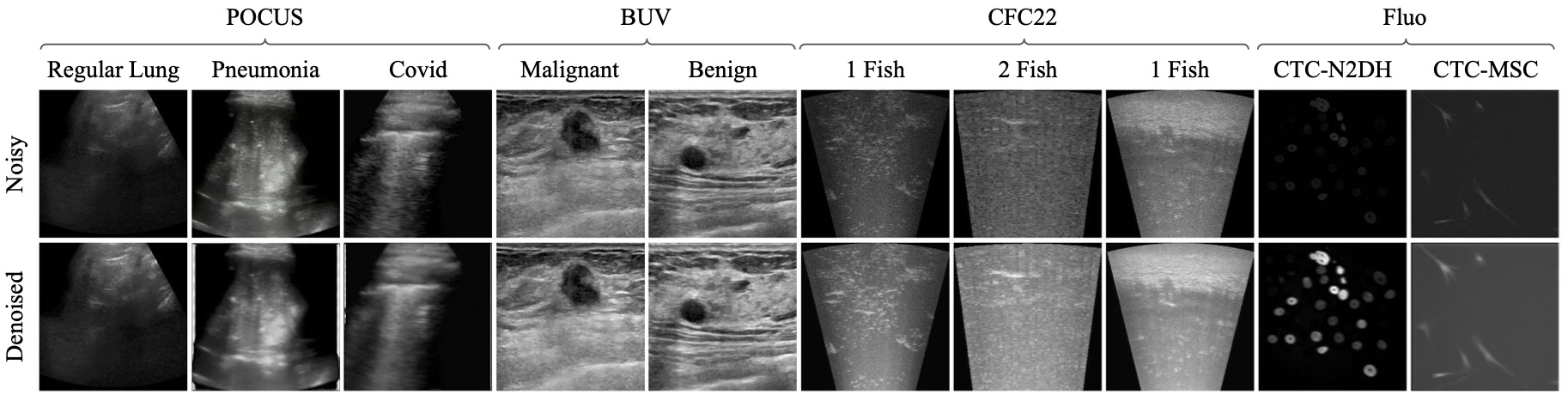}
    \caption{\textbf{Qualitative raw-denoised pairs of \NAME{}}. Qualitative results for \NAME{} trained on POCUS (lung health categorization), BUV (breast lesion detection), CFC22 (fish detection, tracking, and counting), and Fluo (cell denoising).}
    \label{fig:qualitative_pairs}
\end{figure*}

\subsection{Evaluation procedure -- Downstream Tasks}
Given that none of our videos have clean (noise-free) versions, we use the downstream performance tasks' metrics as proxies for our denoised performance. We use $FBD$ from \cref{sec:denoising_metric} when detection annotations are available.

\textbf{Denoising for Detection, Tracking, and Counting.} For CFC22, which has detection, tracking, and counting as downstream tasks, we follow a simplified version of the detection pipeline from \citet{cfc2022eccv} -- we train a YOLOv5 model for 5 epochs with the longest side of an image set to 896 and no augmentations. We remove duplicate predictions using non-maximal suppression. We use mAP$_{50}$ \cite{map50} to evaluate detection performance frame-by-frame. We use a pretrained-frozen ByteTrack tracker and calculate MOTA \cite{clear}, HOTA \cite{HOTA}, and IDF1 \cite{IDF1} scores for evaluation. More details and hyperparameter settings are in the Supplemental Materials Sec. \ref{sec:cfc_detector_hyperparams} and \ref{sec:cfc_tracker_hyperparams}.  
For counting, we use trajectories from the tracks to create nMAE scores, defined in \citet{cfc2022eccv}, for each domain. The tracking and counting pipelines are training-free.

For BUV, we follow the training procedure of \citet{CVANet}; we also follow their final fine-tuning step and evaluation to generate an AP$_{50}$ metric. Note that we know that breast lesions are darker (rather than brighter) spots in ultrasounds. As a result, we invert our reconstruction error to take the minimum difference rather than the maximum: 
\begin{equation}
\text{inv}S_{t,T} = \min(0, I_{t+T} - I_t) + I_t + \min(0, I_{t-T} - I_t)
\end{equation}
\noindent{}\textbf{Denoising for Classification.} 
For POCUS, we perform 5-fold cross-validation following \citet{born2021accelerating, born2021l2}, ensuring all frames from a video remain in the same fold. We adopt their fine-tuning strategy and hyperparameters. For evaluation, we compute per-class precision, recall, and F1, then average across folds to obtain overall metrics.

\section{Results}\label{sec:results}
\NAME{} boost signal of objects of interest in low signal-to-noise video. It improves a range of downstream tasks in a way that is computationally less resource-intensive and yields higher performance than other denoising methods.

\begin{table}[tp]
    \centering
    \vspace{0pt} 
    \resizebox{0.4\textwidth}{!}{%
    \vspace{0pt} 
\begin{tabular}{l c c c c}
\toprule
 \textbf{Train-Time (hours)} & CFC22 & POCUS & BUV \\
\hline
N2V \cite{krull2019noise2void} &  144 & 0.75 & 1.5\\
UDVD \cite{Sheth_2021_ICCV} & 96* & 12 & 23 \\
UMVD \cite{UMVD} & 91* & 34 & 42\\
LG-BPN \cite{wang2023lgbpn} & 101 & 6 & 4 \\
\NAME{} (Ours) & \textbf{20} & \textbf{0.5} & \textbf{1}\\
\end{tabular}  
    }   
    \caption{\textbf{\NAME{} is resource-efficient during training.} Note that UDVD and UMVD (U*VD) took 8 days to train CFC22, but U*VD trained CFC22 only for one epoch. For all other datasets, U*VD both trained for 10 epochs on 2 NVIDIA RTX 4090 GPUs.}
    \label{tab:denoiser_time}
\end{table}

\subsection{Denoising Performance.} 
\begin{figure}[tp]
    \centering
    \includegraphics[width=0.75\linewidth]{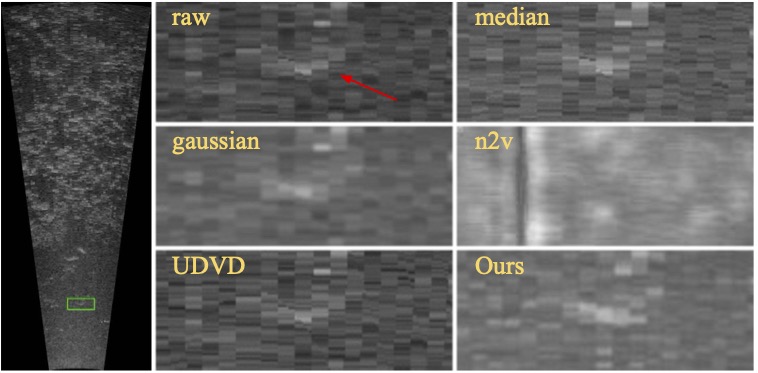}
    \caption{\textbf{Qualitative denoising performance on CFC22}. We can see that the fish is easiest to spot as a bright patch after processing with our denoiser. The green box highlights the fish location. Each denoised image zooms in to that green bounding box. The red arrow in the raw frame points to the fish location. Additional example visualizations are in \cref{fig:additional_fish_denoised} in Supp. Mat.}
    \label{fig:fish_qual_denoising}
\end{figure}

\NAME{} produces clear contiguous objects, where other methods do not, shown in \cref{fig:qualitative_pairs} \& \ref{fig:fish_qual_denoising}. \NAME{} is also able train to convergence in 22\% of the time of the next-quickest network-based method -- more can be seen in \cref{tab:denoiser_time}.

\noindent{}\textbf{Fish Denoising: CFC22.} \NAME{} increases the contrast between fish and background (\cref{fig:fish_qual_denoising}). As such, the distributions of pixel intensities at the same location when fish are present and when they are not are distinct. This is shown in \cref{tab:denoised_metrics_fish}, where \NAME{}'s $FBD$ is significantly higher than that of other methods.
\begin{table}[t]
    \centering
        \resizebox{0.35\textwidth}{!}{%
        \begin{tabular}{lcccc}
\toprule
&\multicolumn{3}{c}{$FBD$ ($\uparrow$})\\
    & Train & Val & Test  \\
    \hline
    Raw & 1005 & 652 & 860 \\
    CFC22++\cite{cfc2022eccv} & \textit{63.7} & 368 & 458  \\
    Median-filtered\cite{gonzalez2006meanfilter} & 523 & 334 & 364 \\
    Gaussian-filtered\cite{gonzalez2006meanfilter} & 793 & 507 & 756 \\
    N2V\cite{krull2019noise2void} & 227 & \textit{194} & \textit{180} \\
    UDVD\cite{Sheth_2021_ICCV} & 402 & 254 & 272 \\
    \NAME{} (Ours) & \textbf{1366} & \textbf{994} & \textbf{1458}\\
    \bottomrule
\end{tabular}    
    }
    \caption{\textbf{\textit{FBD} \cref{eqn:fbd} between P(Fish) vs. Q (Non-fish).} For all ground truth bounding boxes, P and Q are composed as follows: P -- we take the set of pixels in each  box from frames with objects. Q -- we extract the set of pixels from the same box location from a frame where there is no object at that location. \textit{Raw}=raw noisy frame $I_t$, CFC22++\cite{cfc2022eccv} = 3-channel image (raw, background-subtracted, frame-to-frame difference),
    \NAME{}=denoised with motion augmentation, as in \cref{sec:reconstruction}. We calculate the KL-divergence metric, discussed in \cref{sec:KL}. $\uparrow$ indicates the metric is better the larger it is. Best values are \textbf{bolded}, worst values are in \textit{italics.}}
    \label{tab:denoised_metrics_fish}
\end{table}

\noindent{}\textbf{Fluorescent Cells Denoising: Fluo.\ } \NAME{} increases the cells' brightness relative to the background, as seen in \cref{fig:qualitative_pairs}. As is standard\cite{Sheth_2021_ICCV}, and because the data size is small, we only perform qualitative analysis on Fluo.

\subsection{Detection Performance}
\NAME{} outperforms other denoising methods when evaluated on downstream detection tasks of CFC22 and BUV.

\begin{figure}[t]
    \centering
    \includegraphics[width=\linewidth]{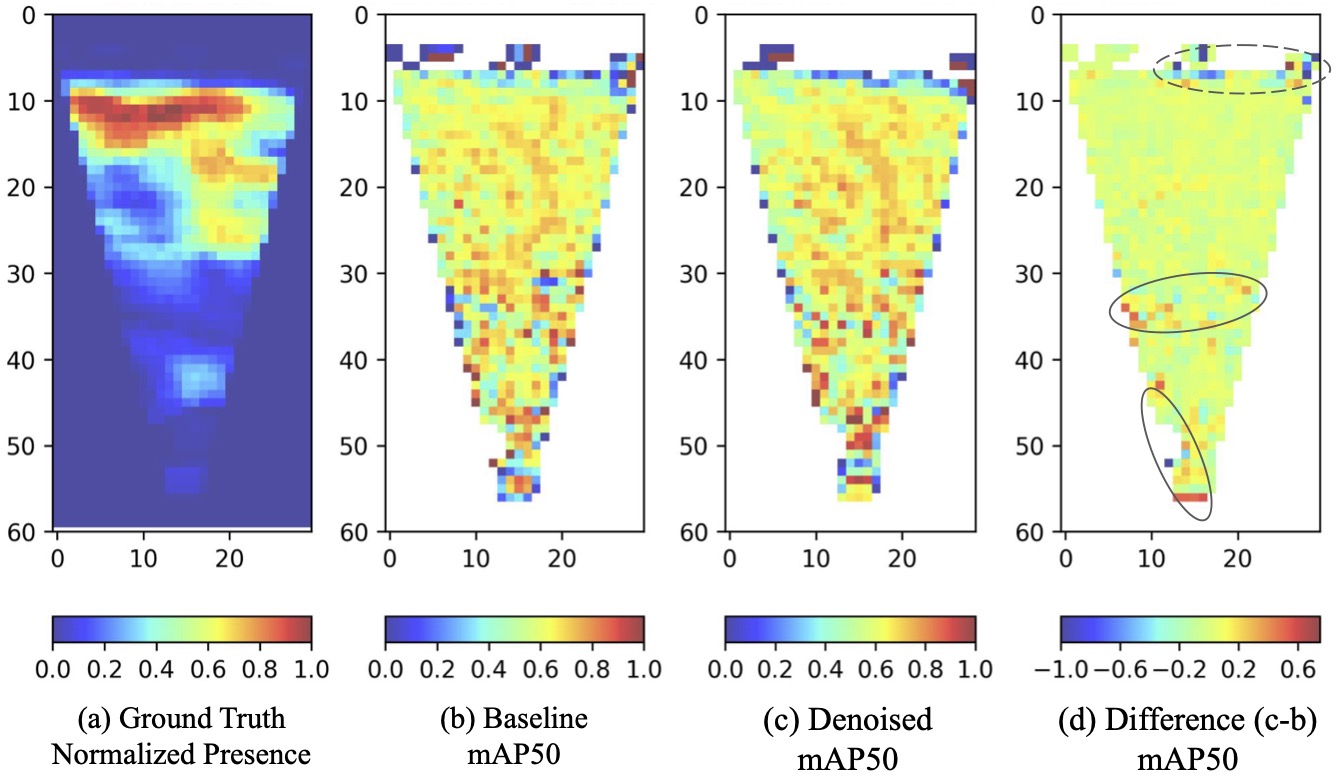}
    \caption{\textbf{Denoising improves detections where signal is infrequent}. \textbf{(a)} the ground truth fish patch locations (from bounding box labels) normalized over the dataset; most fish pass by in the top region, fish crossings below are infrequent, thus there is more training signal in the top part of the videos. \textbf{(b\&c)} patchwise detection performance of CFC22++\cite{cfc2022eccv} and \NAME{}, respectively, on the CFC22 dataset. Heatmaps indicate mAP$_{50}$ performance over all frames of the test set at pixel patches. The more red a patch is, the higher the mAP$_{50}$ of that patch; the more blue the patch is, the lower the mAP$_{50}$. \textbf{(d)} the difference, \NAME{} - CFC22++, with solid ellipses at regions of heightened performance and dashed ellipses around areas of lowered performance. Denoising improves detections in areas where signal is infrequent. On the other hand, detection performance declines in areas where signal is abundant. Additional patch maps can be seen in \cref{fig:trajectory_example_with_detections} in the Supp Mat.}
    \label{fig:heatmaps}
\end{figure}

\noindent{}\textbf{Fish Detection: CFC22.} The detection performance of \NAME{}-processed frames is better than detection performance of other processed frames for CFC22. This is shown in \cref{fig:fish_qual_denoising} \& \ref{fig:heatmaps} and  \cref{tab:big_table}. \NAME{} improves detection performance in areas where objects and signal are rare. \NAME{} frames result in an improvement of 43.2\% and 9.4\% test accuracy compared to the raw and background-subtracted frames respectively, and a 5.1\% boost in performance compared to a three-channel image (raw, background-subtracted, and frame-to-frame-absolute difference) described as baseline++ in \citet{cfc2022eccv}, but hereon referred to as CFC22++. 
\begin{table*}[htp]
    \centering
    \vspace{0pt} 
    \resizebox{0.75\textwidth}{!}{%
    \begin{tabular}{l c c c c c c c }
 \toprule
 & \multicolumn{3}{c}{CFC22 (Test)}& \multicolumn{3}{c}{POCUS (5-fold-CV)} & \multicolumn{1}{c}{BUV(Test)} \\
 \cmidrule(lr){2-4} \cmidrule(lr){5-7} \cmidrule(lr){8-8} 
 Method & mAP$_{50}$\cite{map50}$\uparrow$ & MOTA\cite{clear}$\uparrow$ & nMAE\cite{cfc2022eccv}$\downarrow$ & AP$\uparrow$ & AR$\uparrow$ & F1$\uparrow$ & mAP$_{50}$\cite{map50}$\uparrow$ \\
 \hline
\multicolumn{2}{@{}l}{\textit{Classical}} \\
Baseline  & 73.8 & 37.4 & 54.8 & 82.6 & 82.0 & 80.4 & 46.4\\
Median-Filter\cite{gonzalez2006meanfilter} & 73.7 & 37.8 & 53.0 & 86.2 & 85.5 & 85.3 & 52.4\\
Mean-Filter\cite{gonzalez2006meanfilter} & 76.4 & 44.3 & 41.4 & 84.0 & 84.7 & 83.2 & 52.6 \\
Gaussian-Filter\cite{gonzalez2006meanfilter} & 74.9 & 27.6 & 56.8 & 84.1 & 84.3 & 83.3 & 46.5 \\
\hline                    
\multicolumn{2}{@{}l}{\textit{Deep-Learning-Based Image/Video Denoising/Restoration}}                  \\  
N2V\cite{krull2019noise2void} & 67.2 & 34.2 & 34.3 & 83.7 & 82.7 & 82.4 & 46.6\\
UDVD\cite{Sheth_2021_ICCV} & 67.2 & 28.1 & 41.9 & 83.7 & 84.6 & 83.4 & 49.9 \\
UMVD\cite{UMVD} & 70.4 & 38.2 & 34.0 & 80.4 & 82.6 & 80.7 & 53.9 \\
LG-BPN\cite{wang2023lgbpn} & 72.6 & 41.9 & 34.2 & 31.9 & 34.3 & 21.5 & 49.7 \\
RVRT\cite{liang2022rvrt} & 62.9 & 29.3 & 43.8 & 81.8 & 79.8 & 78.7 & 26.0 \\
AverNet\cite{zhao2024avernet} & 65.4 & 30.3 & 35.7 & 83.7 & 83.4 & 82.6 & 47.0 \\

\hline                   \multicolumn{2}{@{}l}{\textit{DAEs}}          \\ 
AE & 67.8 & 34.3 & 41.7 & 82.3 & 84.7 & 82.1 & 46.9 \\
UNet \cite{unet} & 73.9 & 34.1 & 56.3 & 83.7 & 84.6 & 83.4 & 51.0 \\
UNet3D\cite{unet} & 66.9 & 32.4 & 35.4 & 83.5 & 80.1 & 80.6 & 47.6 \\
\NAME{} (Ours) & \textbf{77.6} & \textbf{47.4} & \textbf{33.9} & \textbf{87.5} & \textbf{86.7} & \textbf{86.3} & \textbf{59.5} \\
\bottomrule
\end{tabular}  
    }
    \caption{\textbf{Downstream results.} \NAME{} does well across all datasets and downstream tasks. Best performance is \textbf{bolded}. Baseline refers to raw for medical ultrasound (POCUS \cite{born2021l2, born2021accelerating} and BUV \cite{CVANet}) and the strengthened baseline CFC22++\cite{cfc2022eccv} for fish sonar (CFC22 \cite{cfc2022eccv}. AP=average precision, AR=average recall, F1=average F1, mAP$_{50}$=mean average precision of detections at IOU threshold 0.5, MOTA=Multi-Object Tracking Accuracy\cite{clear}, nMAE=normalized mean absolute counting error\cite{cfc2022eccv}. More tracking results are in \cref{fig:track}. }
    \label{tab:big_table}
\end{table*}

\noindent{}\textbf{Breast Lesion Detection: BUV.}
\NAME{} clarifies the breast lesion imagery, as seen in \cref{fig:qualitative_pairs} and results in a boost of over 11\% in breast lesion detection mAP$_{50}$ scores compared to the next best method (UMVD). More is in \cref{tab:big_table}.

\subsection{Tracking and Counting Performance}
\begin{figure}
    \centering
    \includegraphics[width=\linewidth]{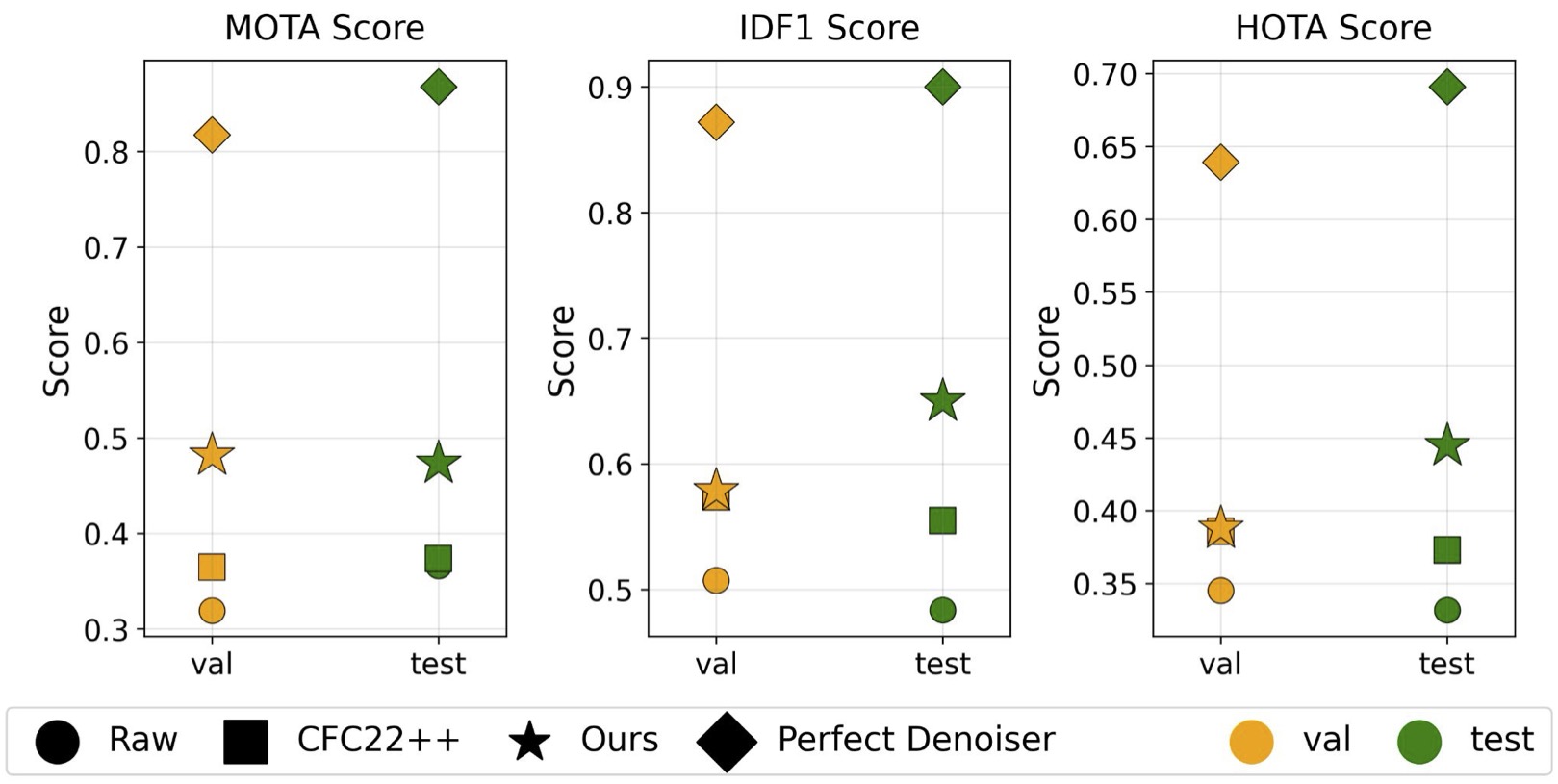}
    \caption{\textbf{Quantitative tracking improvements.} CFC22++ consists of the three-channel (background-subtracted, absolute-difference, raw) frames. The ``Perfect Denoiser'' refers to frames that have black backgrounds and white masks at the bounding box locations. Denoising results in higher MOTA scores for val and test; \NAME{} boosts IDF1 and HOTA scores in test moreso than in val.}
    \label{fig:track}
\end{figure}

\textbf{Fish Tracking and Counting: CFC22.} Compared to classical and other DNN methods, \NAME{} frames yield higher tracking and counting performance (\cref{fig:track}, \cref{tab:big_table}). Stronger fish–background separation reduces false negatives and increases true positives in detections, subsequently improving tracker and count accuracy.


\subsection{Categorization Performance} 
\begin{figure}[tp]
    \centering
    \includegraphics[width=0.75\linewidth]{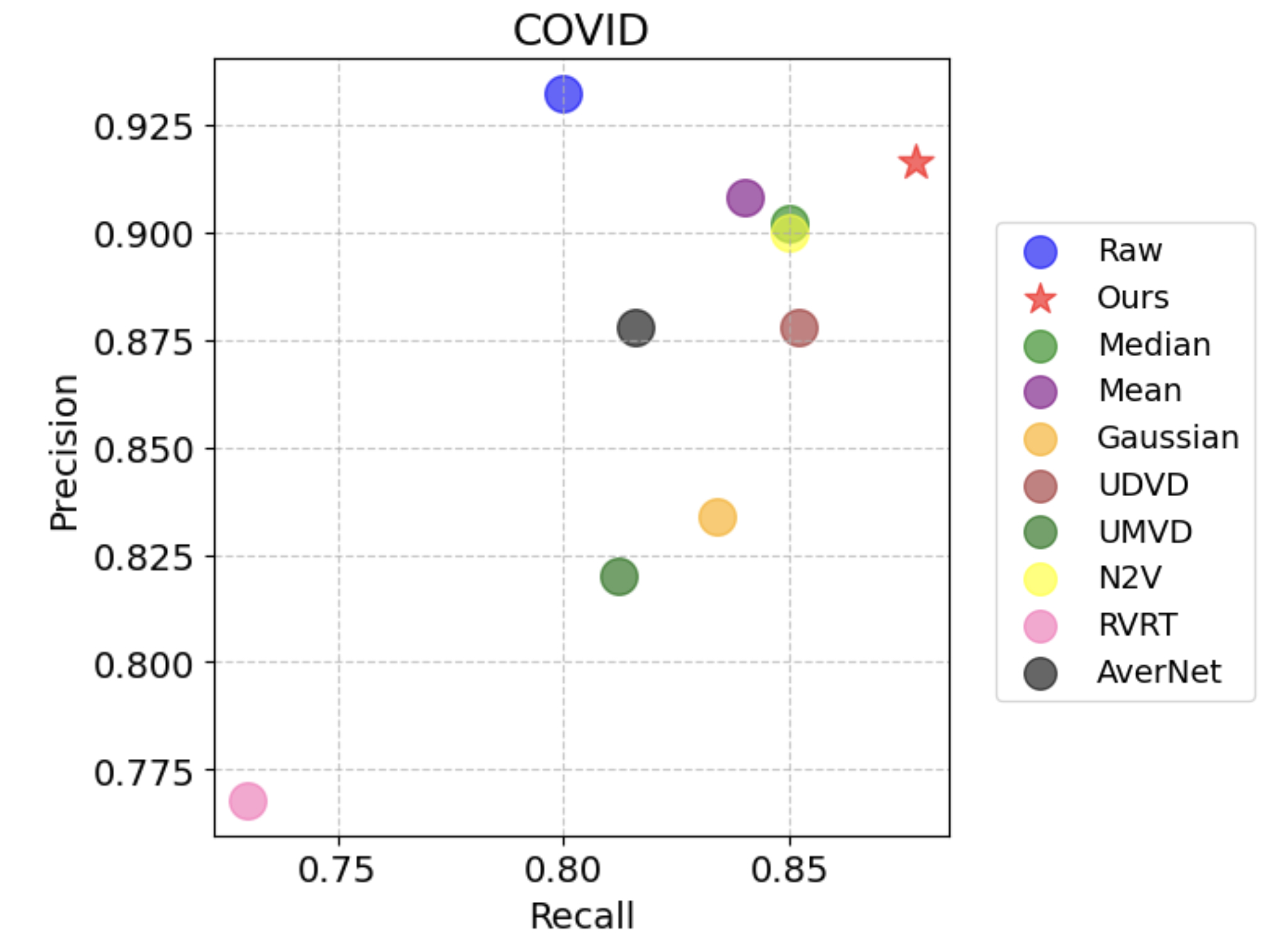}
    \caption{\textbf{Covid Precision-Recall across denoising methods}. \NAME{} has the highest average precision and average recall across denoising methods. Additional class-wise performance comparisons are in \cref{fig:classwise_POCUS} in the Supplementary materials.}
    \label{fig:covid}
\end{figure}

\textbf{Lung Health Categorization: POCUS.}
Images processed through \NAME{} yield the best 5-fold cross-validation image classification score compared to classical and network-based denoising methods on lung categorization, shown in \cref{tab:big_table}. \cref{fig:covid} shows the precision-recall for the denoising methods on the Covid class -- \NAME{} has the highest accuracy for Covid classification. Additional per-class performance analysis is in Sec.\ \ref{sec:pocus_per_class_supp} of the Supp. mat.

\subsection{Ablations}
We ablate the reconstruction target and the denoising autoencoder to find their relative importance.

\noindent{}\textbf{Reconstruction Target}. PFDwT1 is the most effective reconstruction target for increasing the accuracy of downstream tasks. We compared PFDwT1 to PFDwT2, $\sigma$ (the standard deviation over input frames), $\Sigma-5\bar{I}$ (the sum of 5 consecutive frames - 5*mean frame), and $\Sigma-3\bar{I}$ (the sum of 3 consecutive frames - 3*mean frame). The results are shown in  \cref{tab:ablation_reconstruction_targets} and \cref{tab:targets} in the Supplementary material.

\noindent{}\textbf{Autoencoder vs. no Autoencoder}. Using a DAE improves downstream detection performance over using the reconstruction targets alone for all targets. More detail is in \cref{tab:ablation_reconstruction_targets} and \cref{tab:targets} in the Supplementary material.


\noindent{}\textbf{Architectures} Our small architecture has better performance on downstream tasks compared to larger architectures. Comparisons of \NAME{} with a vanilla Autoencoder, a UNet\cite{unet}, and a UNet\cite{unet} with 3D convolutional kernels are shown in \cref{tab:big_table}. For additional details and architectures, see \cref{tab:additional_ablations} in the Supp. mat. The vanilla Autoencoder's architecture is also explicitly defined in \cref{tab:autoencoder_arch}.

\section{Discussion and Conclusion}
While \NAME{} is a clear improvement, we recognize that there are further improvements to be made and research directions to be explored.
\label{sec:discussion}

\noindent{}\textbf{Limitations.}\label{sec:limitations}
As the spatiotemporal component of our method relies an object's location to be overlapping in sequential frames, very fast moving objects may decrease performance on downstream tasks using \NAME{}. On the other hand, if objects are stationary, \NAME{} does not improve performance, though it also should not be detrimental.

We recognize that combining noisy frames adds to the noise of the overall signal rather than removing it. Work \cite{bengio2013denoisingautoencoders} has shown that denoising methods capture clean data's underlying structure. Denoising autoencoders purposefully corrupt training data by adding noise or masking some of the input values \cite{tong2022videomae, he2022MAE}. We rely on the autoencoder to remove noise implicitly by focusing on the largest reconstruction areas to minimize loss. This assumes that the objects of interest are larger than the noise signature. Indeed, we found that detection performance occasionally dipped for very small (\eg $\leq$ 10 pixel) objects, when this assumption is not held.

We observe a change in frame noise when there is an object of interest vs. when there is no object of interest: namely, when there is no object of interest, the whole frame is bright, vs. when an object is in the frame, the object pops and the background becomes dim: ideally, the background would stay dim regardless. An approach to handle this would be adding a component to the loss to focus on background consistency. 

Despite these challenges, \NAME{}-processed frames outperform other denoising methods on downstream tasks while requiring fewer training resources.  

\noindent{}\textbf{Future work.} 
We are interested in training end-to-end: combining the representations from the denoiser and the downstream tasks. 
We are also interested in exploring how to broaden and emphasize the motion signature. 
Finally, we recognize the shared qualities of each of these datasets and also understand that self-supervised methods are data-hungry \cite{li2022supervision, singh2024zero}. As such, one could explore the performance benefit of training on all datasets collectively to learn general low-SNR video properties. 

\noindent{}\textbf{Conclusion.}
We present \NAME{}, a self-supervised denoising method that improves downstream performance in low-SNR videos without requiring clean data. Our approach is general and applicable to a range of low-SNR video tasks and domains. It is based on the confirmed intuition that while there is motion in the foreground and background, motion signatures between foreground and background are distinct, and a simple model can separate them to improve the SNR. We also introduce a new metric, $FBD$,  to capture this relationship. \NAME{} enhances object motion while leveraging autoencoders’ denoising capabilities to boost downstream performance efficiently.

\noindent{}\textbf{Acknowledgments.} 
This work was supported by the Caltech Resnick Sustainability Institute Impact Grant ``Continuous, accurate and cost-effective counting of migrating salmon for conservation and fishery management in the Pacific Northwest.'' We also thank Neehar Kondapaneni, Angela Gao, Sara Beery, Justin Kay, and Laure Delisle for helpful feedback.
{
    \small
    \bibliographystyle{ieeenat_fullname}
    \bibliography{main}
}
\clearpage
\setcounter{page}{1}
\maketitlesupplementary
\appendix



We present additional experimental results as ablations Sec.\ (\ref{sec:ablations}), additional implementation details (Sec.\ \ref{sec:implementation_deets}), and additional visualizations (Sec.\ \ref{sec:visualizations}).

\textbf{Benefits and risks of this technology}. Improving classification, tracking, and counting in sonar and ultrasound videos is useful across medical, ecological, and other fields. Counting fish with sonar allows for a non-invasive way to measure population size, which can then be used for conservation and ecological efforts, for understanding effects of climate change, and for monitoring human fishing behavior for economical reasons. Improving classification in ultrasound videos, too, paves a path for more automated diagnosis. Risks, though, are inherent in both tracking applications and applications of sensitive data. Care must be taken when using these models, so that they are not used blindly without human intervention to make decisions.

\section{Additional Experimental Results}\label{sec:ablations}

\subsection{Additional CFC22 Ablation Results}\label{sec:ablation_additions}

As in the main paper, we evaluate CFC22 on the detection val/test splits, and show results using mAP$_{50}$ across the dataset splits. We look at the effect of bottleneck size in the hourglass network, traditional augmentations, input resolution size, and reconstruction targets on how the trained denoiser affects downstream detection performance. We also look at the effect of downstream task performance when using the reconstruction target alone ($S_{t,T}$) compared with using the learned reconstruction ($\hat{S}_{t,T}$).

\textbf{Bottlenecks size}. For all experiments on CFC22, we use a default input size of 1024hx512w, reconstruction target as PFDwT1, mean-squared error (MSE) loss, and we train the denoiser for 20 epochs. Here the hourglass network remains 2 layers, with the number of input channels as 512, but the number of channels in the middle layer changes. We notice that for training, larger (less-restrictive) bottlenecks yield higher performance. For val and test, though, bottleneck sizes over 64 improve performance, but the differences between 128 and 512 is worse for val and negligible for test. Results can be seen in \cref{tab:bottleneck}.

\textbf{Resolution size}. We vary the input resolution size to train the denoiser and notice higher performance for train and test when higher resolutions are used, seen in \cref{tab:resolution}. We hypothesized that higher resolution size would make the denoiser more stable for downstream detections because higher resolution sizes would mean that removing entire fish (\ie small fish) would be less probable. It is interesting to note that the highest resolution size 2048x1024 for val led to lower detection performance than that of resolution size 1024x512. We note, though, that higher resolutions lead to smaller batch sizes and longer training time.

\textbf{Traditional Augmentations}. We apply salt-and-pepper noise, gaussian-blur, motion-blur, brightness, and erasing from the kornia.Augmentations library. We apply these augmentations when training the denoiser. We do \textit{not} apply these augmentations when training downstream tasks. We found that no traditional augmentations to train the denoiser, though, improve downstream performance. Results can be seen in \cref{tab:augmentations}.

\textbf{Reconstruction Targets}. We experimented with a handful of reconstruction targets: 

\textit{Frame difference}---such as absolute difference ($S_{|d|}=|I_t - I_{t+T}|$) or raw difference ($S_d=I_t - I_{t+T}$)---has been used in other self-supervised works as a spatiotemporal reconstruction target \cite{sun2022bkind}. This works well in video where the movement in the background is less than the foreground movement. For our experiments, we use absolute difference as frame difference.

\textit{Raw frame} ($I_t$) predicts the input (identity) frame alone.

\textit{Background subtraction} (bs) We approximate the background frame,  $\bar{I_v}$, as the mean aggregate of video over time. This is based on the approximation that objects of interest are sparse in terms of space and time. The mean frame is subtracted from every frame in the video ($S_d=(I_{v})_t - \bar{I_v})$. 

\textit{Positive Frame Difference with current frame} (PFDwTN). We discuss this in more detail in the main paper, \cref{sec:reconstruction}. We experimented with T=2 (PFDwT2) and T=1 (PFDwT1), ultimately selecting T=1. 

\textit{Standard Deviation across all frames} ($\sigma$) is taken across all of the frames loaded in a window of continuous frames, $\sigma(I_{t-N}:I_{t+N})$ where 2N+1 is the size of the window. We experimented with N=1 and N=2.

\textit{Sum frames minus N*background} ($\Sigma-N\bar{I_v}$) sums all of the frames in a window size N and takes the positive difference $N*\bar{I_v}$ where $\bar{I_v}$ is the mean frame of all frames in a video: $\max(0, (\sum_t^{T}I_t) - N\bar{I_v})$. We experimented with window sizes N=3 and N=5.

Visualizations of all of these can be seen in \cref{fig:reconstruction_targets_supp}

\begin{table}[t]
\centering
\vspace{0pt} 
\begin{tabular}{lcccc}
 \toprule
 & & \multicolumn{3}{c}{mAP$_{50}$} \\
 \cmidrule(r){3-5}
\multicolumn{1}{c}{Signal Modification} & \multicolumn{1}{c}{AE} & \multicolumn{1}{c}{Train} & \multicolumn{1}{c}{Val} & \multicolumn{1}{c}{Test} \\
\midrule
 \toprule
\multicolumn{5}{l}{\textit{Signal Modification w/o Denoising Network}}\\
Raw ($I_t$) & \xmark & 79.6 & 69.6 & 54.2  \\ 
$\sigma$ & \xmark & 79.8 & 69.4 & 72.5  \\ 
$\Sigma-5\bar{I}$ & \xmark & 78.3 & 67.6 &  71.7 \\ 
PFDwT1 & \xmark & 80.2 & 66.9 & 68.2\\ 
PFDwT2 & \xmark & 81.2 & 68.1 &  63.0 \\ 
\multicolumn{5}{l}{\textit{Signal Modification w/ Denoising Network}}\\
Raw ($I_t$) & \cmark & 81.5  & 68.4 & 73.4 \\
$\sigma$ & \cmark & 82.2 & 70.0 & 73.5  \\
$\Sigma-5\bar{I}$ & \cmark & 79.8 & 68.1 &  71.7 \\
\rowcolor{lightgray}
PFDwT1 & \cmark & \textbf{83.5} & \textbf{70.6} & \textbf{77.6} \\
PFDwT2  & \cmark & 82.2 & 68.5 & 71.4 \\
\bottomrule
\end{tabular}

\caption{\textbf{Effect of Different Motion Enhancements with and without \NAME{}'s Autoencoder Network (AE) on CFC22.} All detectors that leverage the AE have superior performance to those that use only the motion-enhanced target on the test set. The modified signal is used as the reconstruction target for the denoising autencoder when it is present, and is the input signal for the downstream task when the autoencoder is not used. All results are on CNNs with skip connections with resolution 1024 and bottleneck 512.}
\label{tab:ablation_reconstruction_targets}
\end{table}

\captionsetup{font=small}
\begin{figure*}[p]
    \centering
    \begin{subfigure}[b]{0.5\linewidth}
        \centering
        \includegraphics[width=0.95\linewidth]{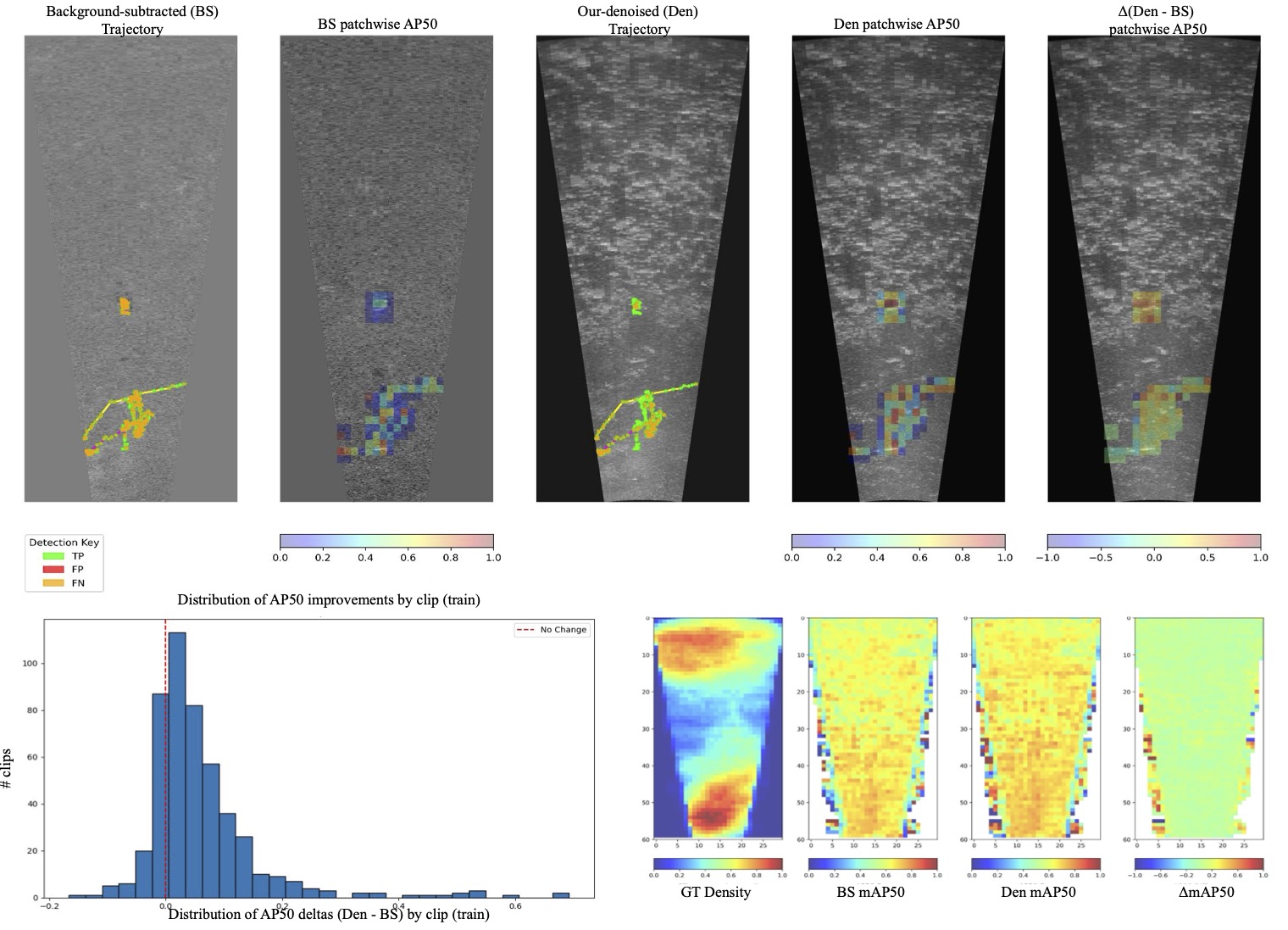}
        \caption{One clip from the CFC22-train river. You can can see the trajectory and patchwise detection performance improves after denoising. Overall, the biggest denoising gains appear to be at the edges of the cone, where fish are known to be small (entering/exiting) but moving.}
        \label{fig:train_trajectory_example_with_detections}
    \end{subfigure}
    \hfill
    \quad
    \begin{subfigure}[b]{0.5\linewidth}
        \centering
        \includegraphics[width=0.95\linewidth]{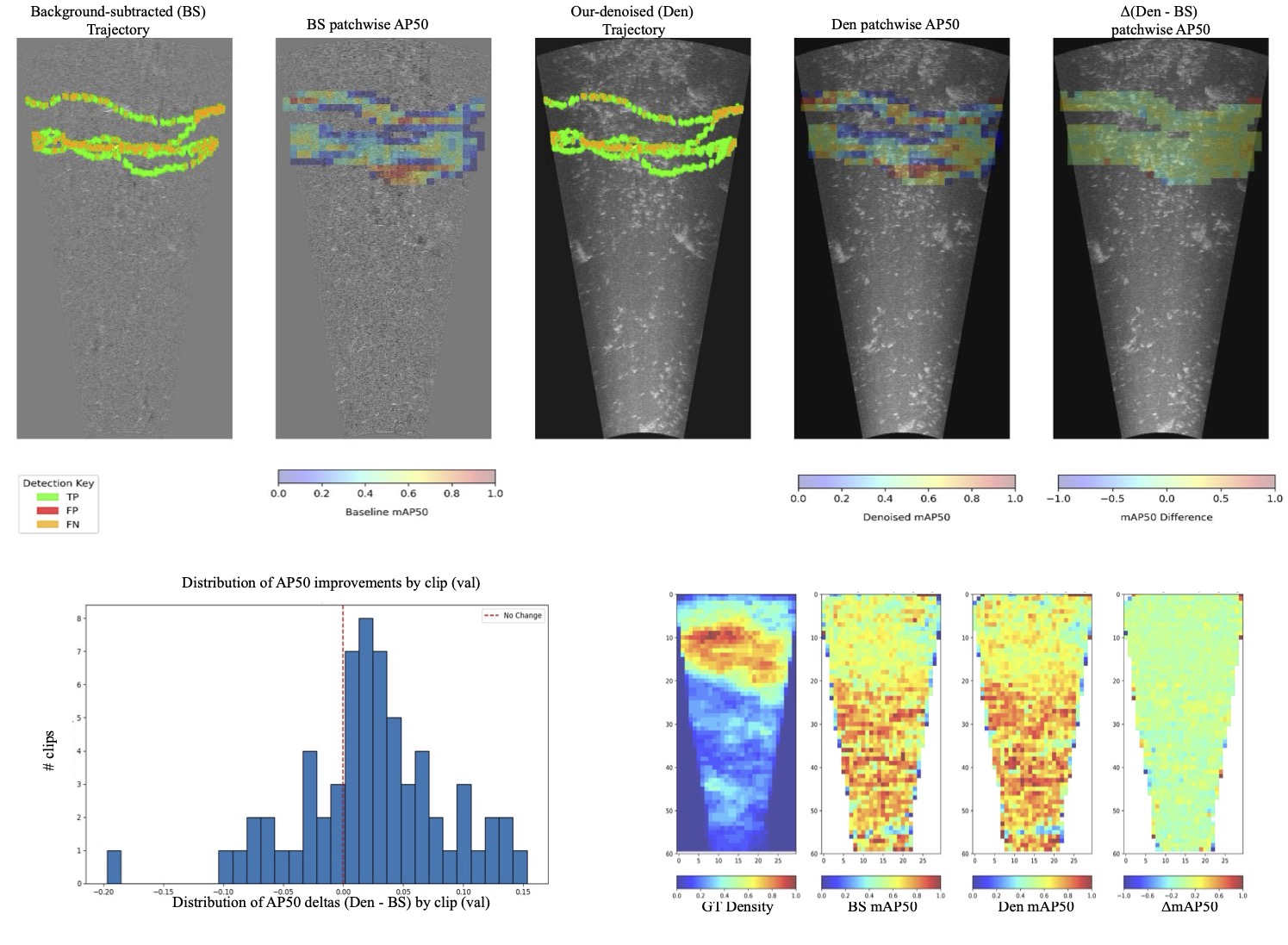}
        \caption{One clip from the CFC22-val river. The denoising gain is smaller and therefore more difficult to see here.}
        \label{fig:val_trajectory_example_with_detections}
    \end{subfigure}
    \hfill
    \quad
    \begin{subfigure}[b]{0.5\linewidth}
        \centering
        \includegraphics[width=0.95\linewidth]{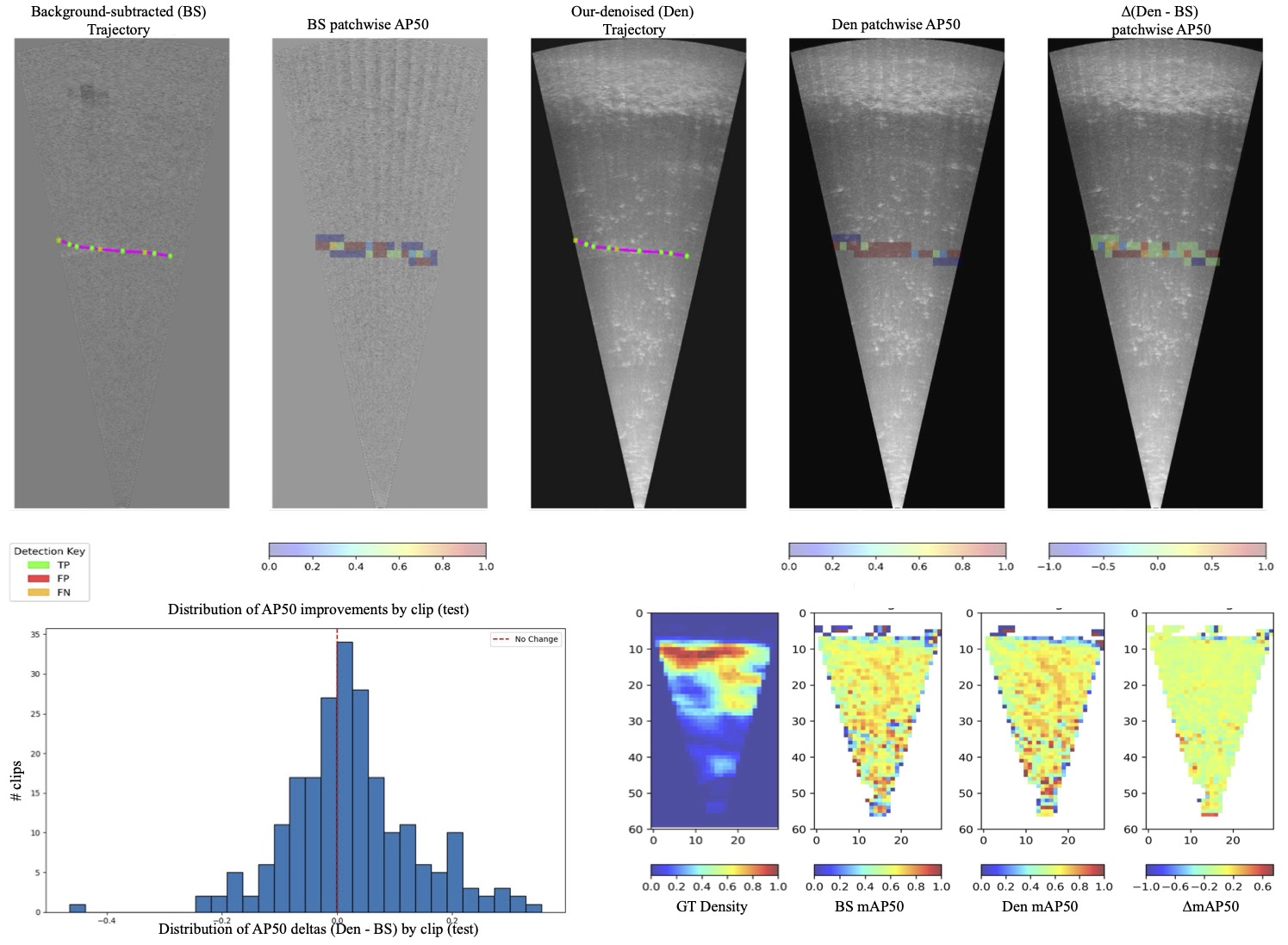}
        \caption{One clip from the CFC22-test river.}
        \label{fig:test_trajectory_example_with_detections}
    \end{subfigure}
    \caption{\textbf{Denoising-improved detection leads to better tracks}. On the single-clip trajectory plots, orange dots indicate false negatives, green dots indicate true positives, red indicates false positives.}
    \label{fig:trajectory_example_with_detections}
\end{figure*}

\captionsetup{font=small}
\begin{figure*}[!htbp]
\centering
\includegraphics[width=0.5\linewidth]{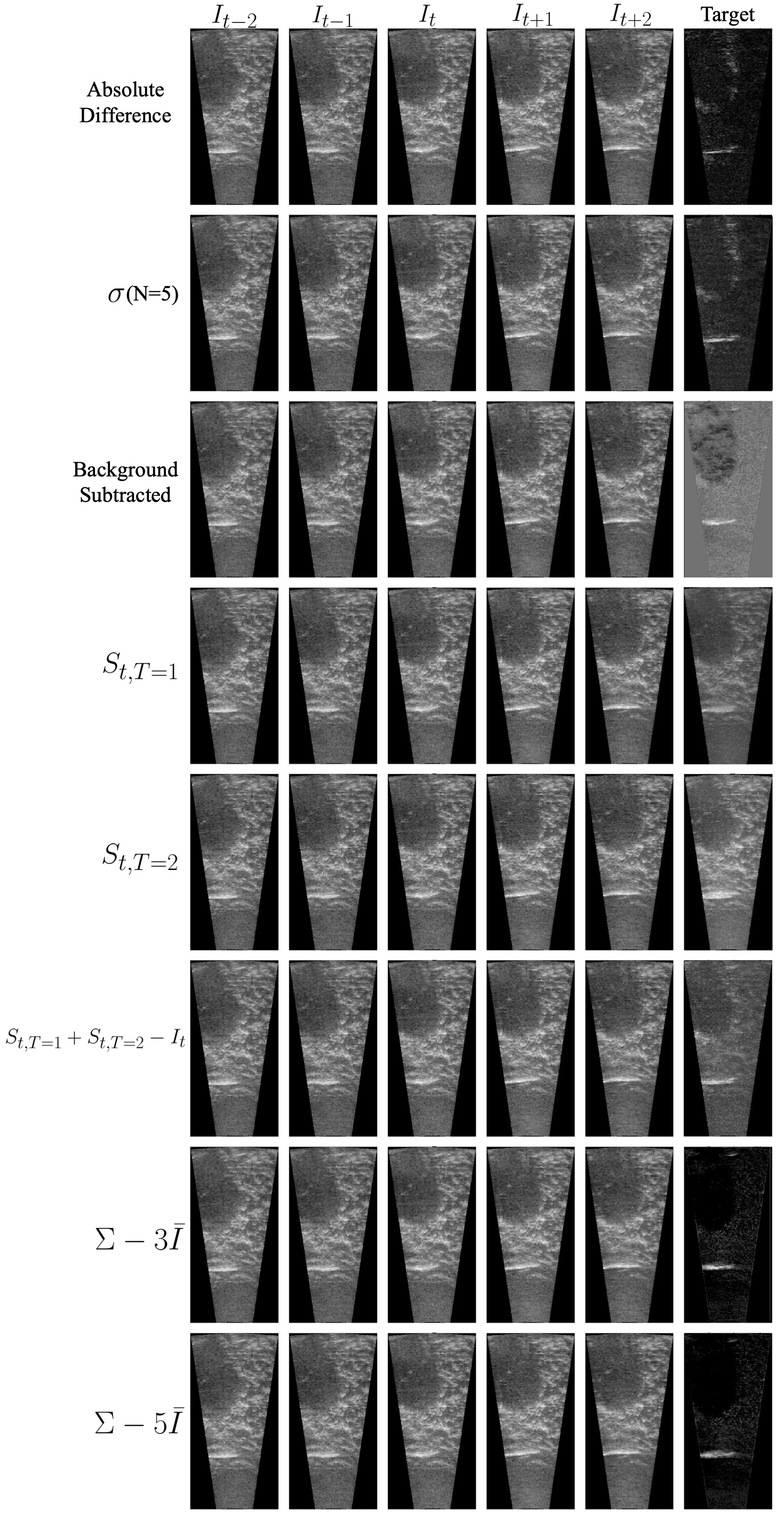}
\caption{\textbf{Reconstruction Targets}. The window T=5 set of frames is shown with each reconstruction target we experimented with on CFC22. While $\Sigma-N\bar{I}$ frames appear strong in this example, we found that empirically they struggled to capture fish that did not move significantly between frames.}
\label{fig:reconstruction_targets_supp}
\end{figure*}

\begin{table*}[bp]
\centering
\scriptsize
\begin{subtable}[t]{0.48\textwidth}
    \centering
    \vspace{0pt} 
    \begin{tabular}{c c c c}
    \toprule
    & \text{Train} & \text{Val} & \text{Test1} \\
    \text{Bottleneck} & \text{mAP$_{50}$} & \text{mAP$_{50}$} & \text{mAP$_{50}$} \\ 
    \hline
    64 & 79.1 & 68.6 & 71.6 \\
    128 & 80.0 & 69.2 & \textbf{72.6} \\ 
    \rowcolor{lightgray}
    512 & \textbf{81.6} & \textbf{69.4} & \textbf{72.6}\\
    \bottomrule
    \end{tabular}
    \caption{\textbf{Bottleneck size.} A larger bottleneck outperforms overly-constricted networks. All results are from CNNs with no skip connections and non-residual blocks.}
    \label{tab:bottleneck}
\end{subtable}\hfill
\quad
\begin{subtable}[t]{0.48\textwidth}
    \centering
    \vspace{0pt} 
    \begin{tabular}{c c c c}
    \toprule
    & \text{Train} & \text{Val} & \text{Test1} \\
    \text{Resolution} & \text{mAP$_{50}$} & \text{mAP$_{50}$} & \text{mAP$_{50}$} \\ 
    \hline
    512 & \textbf{81.3} & \textbf{69.2} & 71.9 \\ 
    \rowcolor{lightgray}
    1024 & 79.1 & 68.6 & 71.6 \\
    2048 & 80.1 & 68.1 & \textbf{72.1} \\ 
    \bottomrule
    \end{tabular}
    \caption{\textbf{Resolution size.} There is no clear optimal - in terms of train and val, the smallest resolution size is the best; however, in terms of test, the largest resolution size is optimal. Note that higher resolutions also lead to longer training times.}
    \label{tab:resolution}
\end{subtable}\hfill
\quad
\begin{subtable}[t]{0.48\textwidth}
    \centering
    \vspace{0pt} 
    \begin{tabular}{l c c c}
    \toprule
    & \text{Train} & \text{Val} & \text{Test1} \\
    \text{Augmentations} & \text{mAP$_{50}$} & \text{mAP$_{50}$} & \text{mAP$_{50}$} \\ 
    \hline
    $\text{saltpepper}_{0.25}$ & 81.2 & 68.5 & 72.2 \\ 
    $\text{saltpepper}_{0.5}$ & \textbf{83.7} & 69.7 & 75.1 \\
    $\text{saltpepper}_{0.75}$ & 81.4 & 69.2 & 72.8 \\ 
    $\text{gaussianblur}_{0.25}$ & 82.1 & 69.9 & 74.8 \\
    $\text{gaussianblur}_{0.5}$ & 81.3 & 68.9 & 75.0 \\
    $\text{gaussianblur}_{0.75}$ & 83.5 & 68.4 & 75.6 \\
    $\text{motionblur}_{0.25}$ & 83.5  & 68.3 & 76.5 \\
    $\text{motionblur}_{0.5}$ & 81.2 & 68.2 & 74.7 \\
    $\text{motionblur}_{0.75}$ & \textbf{83.7} & 69.6 & 73.9\\
    $\text{brightness}_{0.25}$ & \textbf{83.7} & 69.8 & 74.7 \\
    $\text{brightness}_{0.5}$ & 82.2 & 69.0 & 73.9 \\
    $\text{brightness}_{0.75}$ & 83.6 & 69.7 & 76.8 \\
    $\text{erase}_{0.25}$ & 82.0 & 68.7 & 68.0 \\
    $\text{erase}_{0.5}$ & 81.1 & 68.7 & 75.6  \\
    $\text{erase}_{0.75}$ & 77.4 & 59.3 & 62.4 \\
    \rowcolor{lightgray}
    No augmentations & 83.5 & \textbf{70.6} & \textbf{77.6} \\
    \bottomrule
    \end{tabular}
    \caption{\textbf{Augmentations.} Augmentations appear to degrade performace. All augmentation experiments are named as $augmentation_{probability}$. All networks are CNNs with skip connections with resolution 1024 and bottleneck 512.}
    \label{tab:augmentations}
\end{subtable}\hfill
\quad
\begin{subtable}[t]{0.48\textwidth}
    \centering
    \vspace{0pt} 
    \begin{tabular}{l c c c}
    \toprule
    & \text{Train} & \text{Val} & \text{Test1} \\
    \text{Target} & \text{mAP$_{50}$} & \text{mAP$_{50}$} & \text{mAP$_{50}$} \\ 
    \hline
    Raw* & 81.5 & 68.4 & 73.4 \\
    Absolute Difference $|I_t - I_{t+1}|$ & 81.6 & 69.2 & 73.5 \\ 
    Sigma(N=5) & 78.8 & 69.2 & 72.8 \\
    \rowcolor{lightgray}
    $\hat{S}_{t,T=1}$* & 82.7 & 70.0 & 74.0 \\
    $\hat{S}_{t,T=2}$* & 82.8 & \textbf{70.6} & 73.0\\
    $\hat{S}_{t,T=2} + \hat{S}_{t,T=1} - I_t$* & \textbf{83.7} & 69.2 & \textbf{74.6} \\
    $\Sigma-3\bar{I}$ & 80.3 & 68.3 & 69.0\\
    $\Sigma-5\bar{I}$ & 80.7 & 68.7 & 72.0 \\ 

    \bottomrule
    \end{tabular}
    \caption{\textbf{Reconstruction targets.} Reconstruction targets including both the original frame and the next or previous frames do better than reconstruction targets incorporating information from just one. Reconstruction targets with the current frame in have *. All results are on CNNs with resolution 1024 and bottleneck 512 with no skip connection.}
    \label{tab:targets}
\end{subtable}\hfill

\begin{subtable}[t]{0.48\textwidth}
    \centering
    \vspace{0pt} 
    \begin{tabular}{lccc}
    \toprule
    & \text{Train} & \text{Val} & \text{Test1} \\
    \text{Architectures} & \text{mAP$_{50}$} & \text{mAP$_{50}$} & \text{mAP$_{50}$} \\ 
    \hline
    Autoencoder & 82.6 & 68.9 & 67.8 \\
    CNN-fine & 82.7 & 69.1 & 74.0 \\
    \rowcolor{lightgray}
    CNN-SKIP & \textbf{83.5} & \textbf{70.6} & \textbf{77.6} \\ 
    CNN-residual & \textbf{83.5} & 69.2 & 73.1 \\
    CNN-resnet-block & 79.8 & 70.0 & 73.6\\
    UNet-downscaled & 82.1 & 69.1 & 75.8 \\
    UNet & 81.2 & 70.0 & 73.9 \\
    UNet3D & 79.0 & 67.0 & 66.9\\
    \bottomrule
    \end{tabular}
    \caption{\textbf{Denoising backbone architecture.} All experiments have our target from equation 2 ($\hat{S}_{t,T=1}$) as their target. Networks are ordered from smallest (in terms of parameters and TFLOPs) to largest -- it is interesting to note that as model size increases, performance does not necessarily increase. We see the top performer is the CNN-SKIP architecture.}
    \label{tab:architecture}
\end{subtable}\hfill

\caption{\textbf{Additional denoise-detection ablations on CFC22.} All values are generated via the detection stage of our pipeline. All reconstruction targets are sized 1024 x 512 unless otherwise stated. We report the mAP$_{50}$ of the \emph{combined} background-subtracted and target reconstruction frame unless otherwise noted. Default settings are marked in \colorbox{lightgray}{gray}. }
\label{tab:additional_ablations}
\end{table*}
\vspace{1em}

\subsection{POCUS Per-Class Performance}\label{sec:pocus_per_class_supp}
\begin{figure}[tbh]
    \centering
    \includegraphics[width=\linewidth]{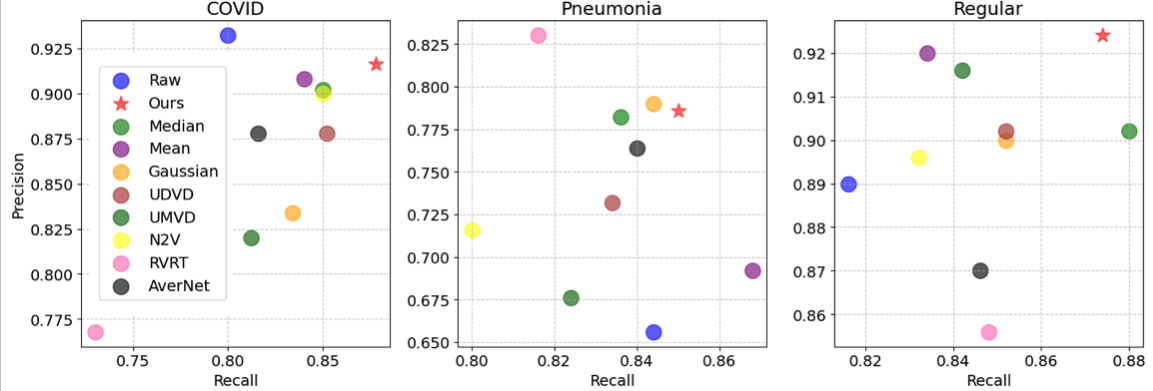}
    \caption{\NAME{} (starred) has high precision and high recall across all POCUS classes.}
    \label{fig:classwise_POCUS}
\end{figure}

\NAME{} performs well across all classes (COVID, Pneumonia, and Regular) in the POCUS dataset (\cref{fig:classwise_POCUS}). For Pneumonia, precision levels across all methods were lower than for other classes. Pneumonia false negatives are more often categorized as Regular than they are Covid across all denoising methods.


\section{Implementation Details}\label{sec:implementation_deets}

\subsection{\NAME{} Architecture Details}
Our method uses a series of convolution blocks with skip connections as an encoder $\Phi$, a bottleneck (hourglass network) $\Theta$, and a reconstruction decoder $\Psi$. Architectural details about each of these are shown in \cref{tab:denoiser_arch}. For more implementation details, the code is publicly available \href{https://github.com/suzanne-stathatos/SAVeD}{here}.

\begin{table*}[t]
    \centering
    \vspace{0pt} 
\begin{tabular}{l c c}
\toprule
\multicolumn{3}{c}{\textit{Encoder}}\\
    Type & Input shape & Output shape\\
    \hline\\
    \text{Conv\_block} & (1,1024,512) & (16, 1024, 512)\\
    Pooling & (16, 1024, 512) & (16, 512, 256) \\
    Skip & (16, 1024, 512) & (16, 512, 256) \\
    \text{Conv\_block} & (16, 512, 256) & (32, 512, 256) \\
    Pooling & (32, 512, 256) & (32, 256, 128) \\
    Skip & (32, 512, 256) & (32, 256, 128) \\
    \text{Conv\_block} & (32, 256, 128) & (64, 156, 128)\\
    Pooling & (64, 156, 128) & (64, 128, 64) \\
    Skip & (64, 156, 128) & (64, 128, 64) \\
    \text{Conv\_block} & (64, 128, 64) & (128, 128, 64)\\
    Pooling & (128, 128, 64) & (128, 64, 32)\\
    Skip & (128, 128, 64) & (128, 64, 32)\\
    \text{Conv\_block} & (128, 64, 32) & (256, 64, 32) \\
    Pooling & (256, 64, 32) & (256, 32, 16) \\
    Skip & (256, 64, 32) & (256, 32, 16) \\
    \text{Conv\_block} & (256, 32, 16) & (512, 32, 16) \\
    Pooling & (512, 32, 16) & (512, 16, 8) \\
    Skip & (512, 32, 16) & (512, 16, 8) \\    
\bottomrule
\toprule
\multicolumn{3}{c}{\textit{Decoder}}\\
    Type & Input shape & Output shape\\
    \hline\\
    Upsample\_block & (512, 16, 8) & (256, 32, 16) \\
    Skip\_connect & (256, 32, 16) & (768, 32, 16) \\
    Conv\_block & (768, 32, 16) & (512, 32, 16)\\
    Upsample\_block & (512, 32, 16) & (256, 64, 32) \\
    Skip\_connect & (256, 64, 32) & (512, 64, 32)\\
    Conv\_block & (512, 64, 32) & (256, 64, 32) \\
    Upsample\_block & (256, 64, 32) & (128, 128, 64) \\
    Skip\_connect & (128, 128, 64) & (256, 128, 64) \\
    Conv\_block & (256, 128, 64) & (128, 128, 64) \\
    Upsample\_block & (128, 128, 64) & (64, 256, 128) \\
    Skip\_connect & (64, 256, 128) & (128, 256, 128) \\
    Conv\_block & (128, 256, 128) & (64, 256, 128) \\
    Upsample\_block & (64, 256, 128) & (32, 512, 256) \\
    Skip\_connect & (32, 512, 256) & (64, 512, 256) \\
    Conv\_block & (64, 512, 256) & (32, 512, 256) \\
    Upsample\_block & (32, 512, 256) & (1, 1025, 512) \\
\bottomrule
\end{tabular}    
    \caption{\textbf{Architecture details of the encoder, bottleneck, and decoder of \NAME{}.} ``\text{Conv\_block}" is a basic convolutional block composed of 3x3 convolution with padding side of 1 and ReLU activation. ``Skip" is a skip connection (stored to be input into the decoder) composed by maxpooling and then running a 1x1 convolution. ``Upsample\_block" is a 2D ConvTranspose with a 2x2 kernel and a stride of 2 and a ReLU activation. ``Skip\_connect'' is the concatenation of the output from Upsample\_block+Conv\_block and the ``Skip'' corresponding to the same layer saved by the encoder. Note that this architecture is on input size of 1024x512.}
    \label{tab:denoiser_arch}
\end{table*}

\begin{table*}[t]
    \centering
    \vspace{0pt} 
\begin{tabular}{l c c}
\toprule
\multicolumn{3}{c}{\textit{Encoder}}\\
    Type & Input shape & Output shape\\
    \hline\\
    \text{Conv\_block} & (3, 1024, 512) & (16, 1024, 512)\\
    Pooling & (16, 1024, 512) & (16, 512, 256) \\
    \text{Conv\_block} & (16, 512, 256) & (32, 512, 256) \\
    Pooling & (32, 512, 256) & (32, 256, 128) \\
    \text{Conv\_block} & (32, 256, 128) & (64, 156, 128)\\
    Pooling & (64, 156, 128) & (64, 128, 64) \\
    \text{Conv\_block} & (64, 128, 64) & (128, 128, 64)\\
    Pooling & (128, 128, 64) & (128, 64, 32)\\
    \text{Conv\_block} & (128, 64, 32) & (256, 64, 32) \\
    Pooling & (256, 64, 32) & (256, 32, 16) \\
    \text{Conv\_block} & (256, 32, 16) & (512, 32, 16) \\
    Pooling & (512, 32, 16) & (512, 16, 8) \\
\bottomrule
\toprule
\multicolumn{3}{c}{\textit{Decoder}}\\
    Type & Input shape & Output shape\\
    \hline\\
    Bilinear\_upsample\_block & (512, 16, 8) & (512, 32, 16) \\
    Conv\_block & (512, 32, 16) & (256, 32, 16)\\
    Bilinear\_upsample\_block & (256, 32, 16) & (256, 64, 32) \\
    Conv\_block & (256, 64, 32) & (128, 64, 32) \\
    Bilinear\_upsample\_block & (128, 64, 32) & (128, 128, 64) \\
    Conv\_block & (128, 128, 64) & (64, 128, 64) \\
    Bilinear\_upsample\_block & (64, 128, 64) & (64, 256, 128) \\
    Conv\_block & (64, 256, 128) & (32, 256, 128) \\
    Bilinear\_upsample\_block & (32, 256, 128) & (32, 512, 256) \\
    Conv\_block & (32, 512, 256) & (16, 512, 256) \\
    Bilinear\_upsample\_block & (16, 512, 256) & (16, 1024, 512) \\
    Conv\_block & (16, 1024, 512) & (1, 1024, 512)\\
\bottomrule
\end{tabular}    
    \caption{\textbf{Architecture details of the vanilla autoencoder.} ``\text{Conv\_block}" is a basic convolutional block composed of 3x3 convolution with padding side of 1 and ReLU activation.  ``Bilinear\_upsample\_block" is a Bilinear Upsample kernel with a scale factor of 2 and align corners set to True. Note that this architecture is on input size of 1024x512.}
    \label{tab:autoencoder_arch}
\end{table*}

\subsection{\NAME{} Hyperparameter Comparisons}\label{sec:our_denoiser_hyperparams_and_details}
The hyperparameters for our method are in \cref{tab:hyperparams}. All DAE models are trained until the training loss converges on 2 NVIDIA RTX 4090 GPUs. 

\begin{table*}[t]
    \centering
    \resizebox{\textwidth}{!}{%
        \vspace{0pt} 
\begin{tabular}{l c c c c c c c c}
\toprule
Dataset & Resolution & Target & Epochs & Batch size & Learning Rate & Optimizer & Scheduler \\
\hline
CFC22 & (1024,512) & $S_{t,T=1}$ & 20 & 16 & 0.0005 & AdamW & Plateau f=0.1 pat=2 \\
POCUS & (1024,512) & $S_{t,T=1}$ & 120 & 8 & 0.0005 & AdamW & Step ss=2, $\gamma=0.05$\\
BUV & (1024,1024) & ${\text{inverse}(S_{t,T=1})}$ & 40 & 8 & 0.0005 & AdamW & Step ss=2, $\gamma=0.05$\\
Fluo & (1024,1024) & $I_t$ & 1000 & 8 & 0.0005 & AdamW & Step ss=2, $\gamma=0.05$\\
\bottomrule
\end{tabular}
    }
    \caption{\textbf{\NAME{} Hyperparameters.} Note ``$\text{inverse}(S_{t,T=1})$''$=\min(0, I_t - I_{t-T}) + I_t + \min(0, I_t - I_{t+T})$. f=Factor, pat=Patience, ss=Step size.}
    \label{tab:hyperparams}
\end{table*}


\begin{figure}
    \centering
    \includegraphics[width=\linewidth]{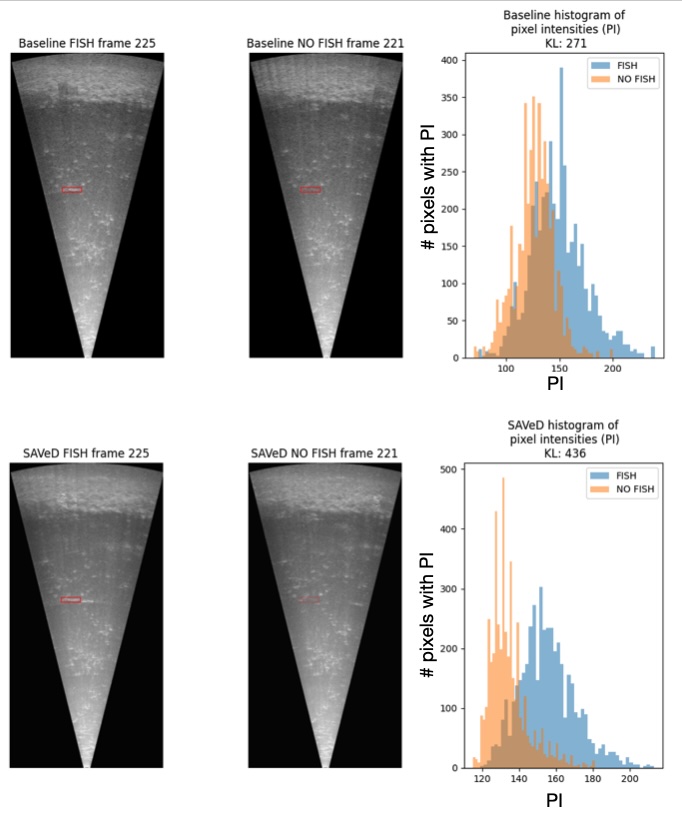}
    \caption{\textbf{Visualization of $FBD$}. Both images on the left are noisy images. The image on the far left has a fish located in the red bounding box. The image in the middle is a frame from the same video clip but with no fish in the red box. The histogram compares the pixel intensity values of the pixels within the bounding boxes. We can see these distributions, while overlapping, are distinct.}
    \label{fig:denoiser_metric_details}
\end{figure}

\subsection{CFC22 Detector Details}\label{sec:cfc_detector_hyperparams}
We fine-tune a YoloV5-small model pretrained on COCO using the default training settings from Ultralytics over 5 epochs with a batch size of 16. As in \citet{cfc2022eccv}, we resize all inputs to have 896 pixels as their longest side; the learning rate is 0.0025. We select the best model checkpoint based on validation mAP$_{50}$. We train on two NVIDIA RTX A6000 GPUs. We recognize that the number of epochs (5) differs from the number of epochs in the original paper (150), and that is intentional. The reasoning is two-fold: 1.) CFC22++ Val and Test Performance after 5 epochs are $<1\%$ lower than Val and Test Performance after 150 epochs, therefore our denoised improvement beats the CFC22++ method also after CFC22++ is trained for 150 epochs while the detector model based on \NAME{} frames is trained for 5 epochs; 2.) We wanted to show that a very simple detector could be used as a result of passing in denoised frames.

\begin{figure}[tp]
    \centering
    \includegraphics[width=\linewidth]{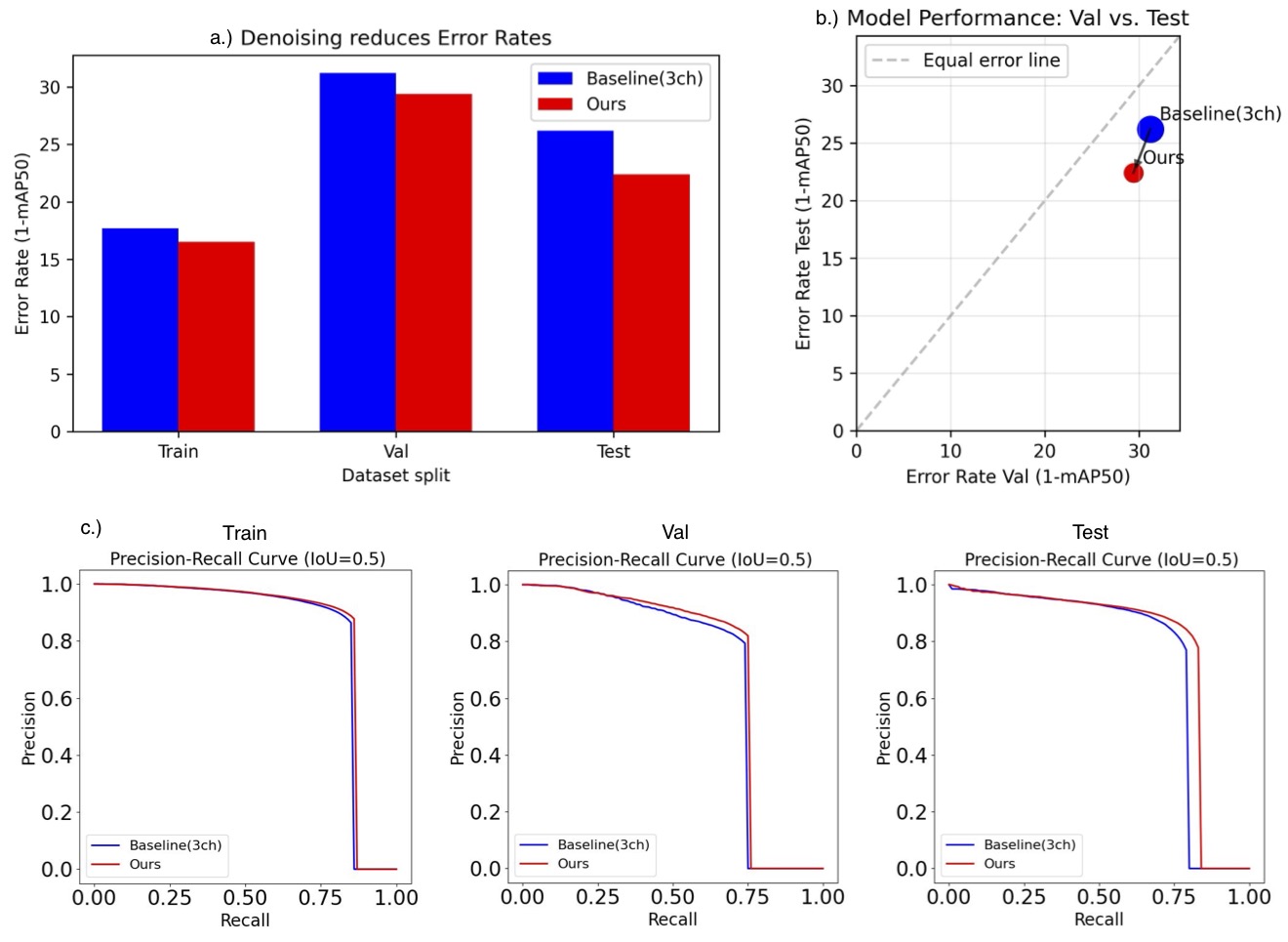}
    \caption{\textbf{Denoising lowers detections error-rates by improving precision and recall} (a) shows baseline detection error (1-mAP$_{50}$) compared to our detection error after our denoising preprocessing step. For all splits train, val, and test, denoising results in lower error. (b) compares error rates from the validation set (x-axis) to error rates from the test set (y-axis) to see how denoising impacts each split. There is a 5.8\% reduction in error in the val set and a 14.5\% reduction in error on the test set. (c) Shows inverted Precision-Recall plots for each CFC22 dataset split -- precision and recall both improve for all splits.}
    \label{fig:fig5}
    \vspace{1em} 

    \includegraphics[width=\linewidth]{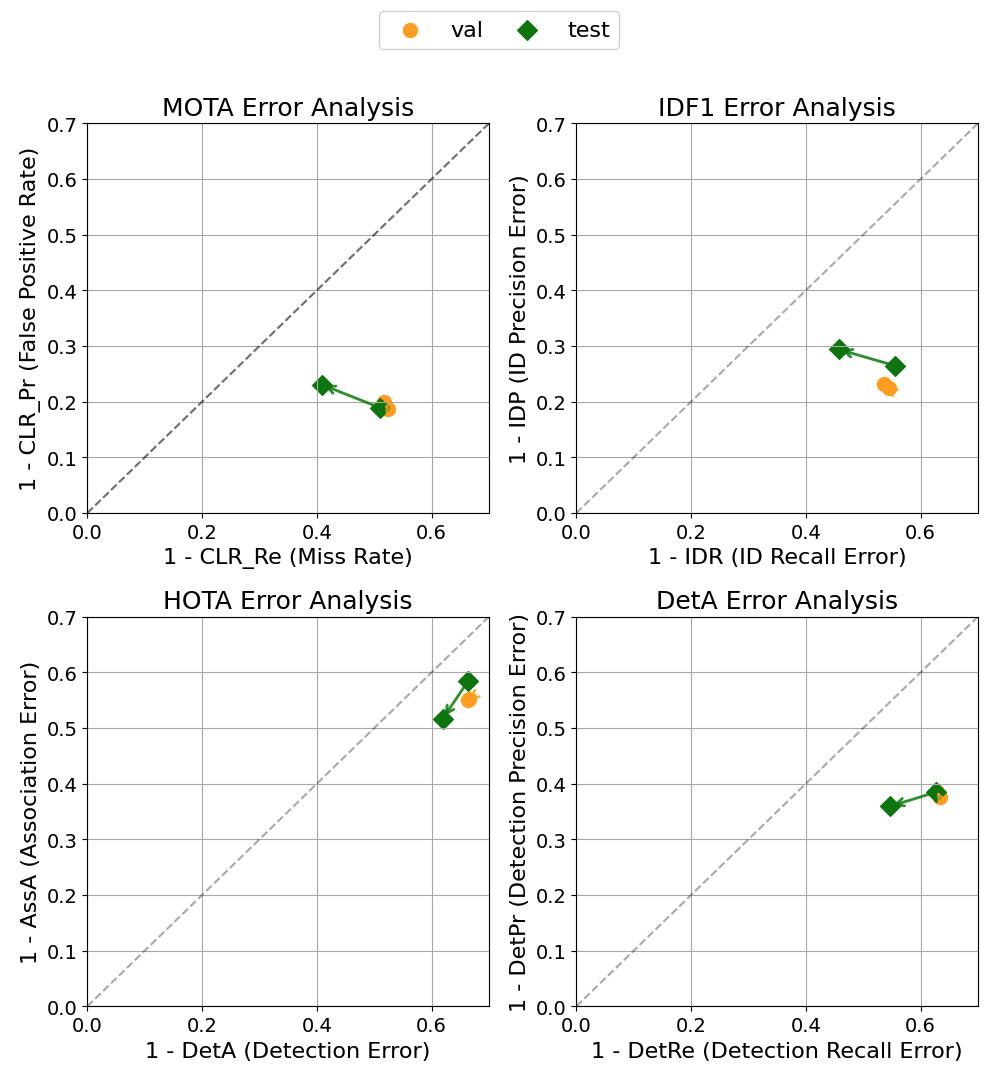}
    \caption{\textbf{Breakdown of track performance improvements for CFC22 val and test}. We can see test improves far more than val, as is standard for the CFC22 dataset.}
    \label{fig:track_perf_deepdive}
\end{figure}

\subsection{CFC22 Tracker Details}\label{sec:cfc_tracker_hyperparams}
We use a pretrained ByteTrack tracker with hyperparameters selected as the optimal hyperparameters for tracking performance on the validation set. Max age, the time until a missing or occluded object is assigned a new id, is 20; Min hits, the minimum number of frames with a track for the track to be considered valid, is 11; IOU threshold, the iou required for an object to be considered the same in the subsequent frame, is 0.01.

\section{Visualizations}\label{sec:visualizations}
\begin{figure}[htbp]
    \centering
    \includegraphics[width=\linewidth]{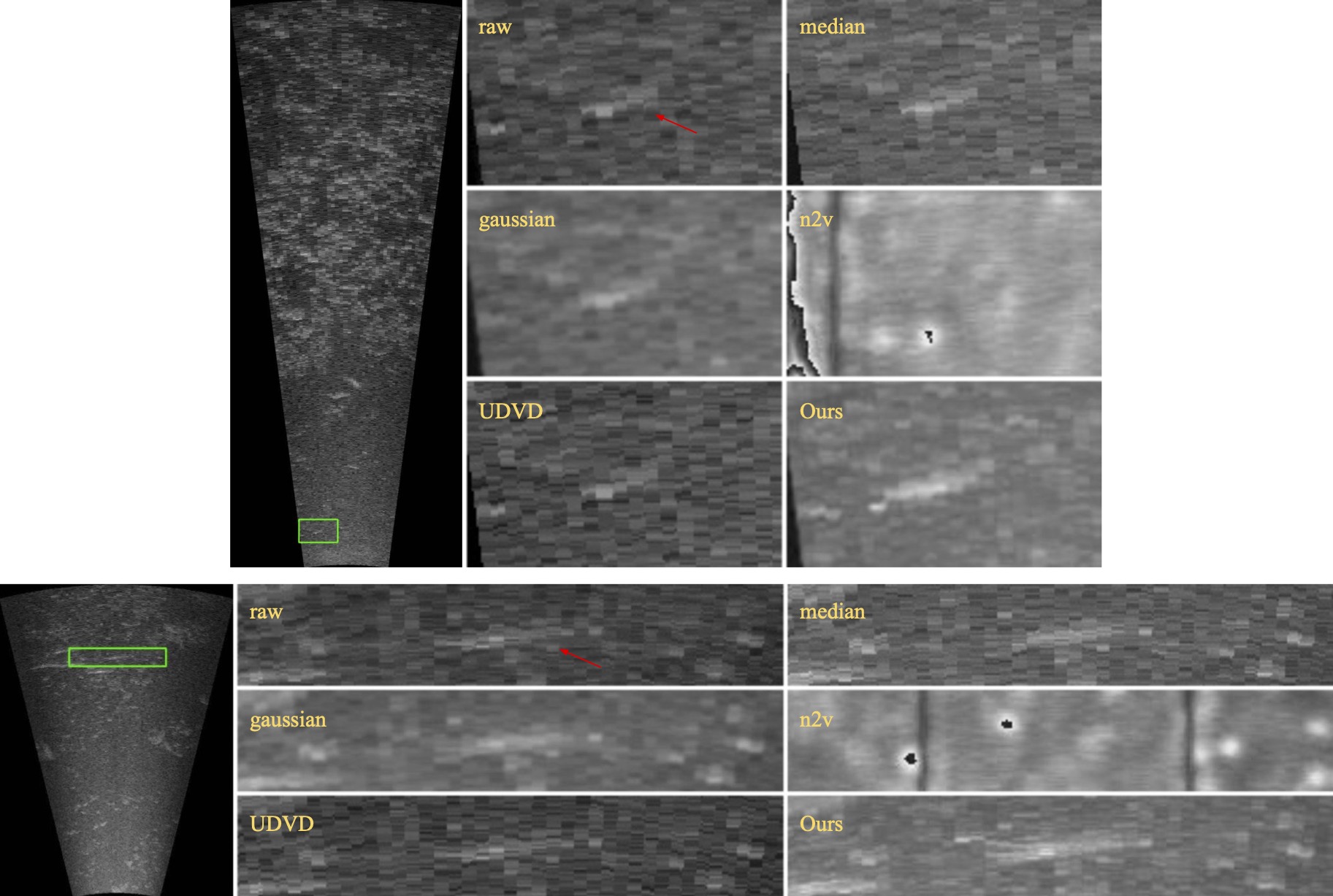}
    \caption{\textbf{Additional visualizations of denoising methods on CFC22}}
    \label{fig:additional_fish_denoised}
\end{figure}

Additional visualizations of the denoising performance on fish in sonar (CFC22\cite{cfc2022eccv})
can be seen in \cref{fig:additional_fish_denoised}).

\begin{figure}[htbp]
    \centering
    \includegraphics[width=\linewidth]{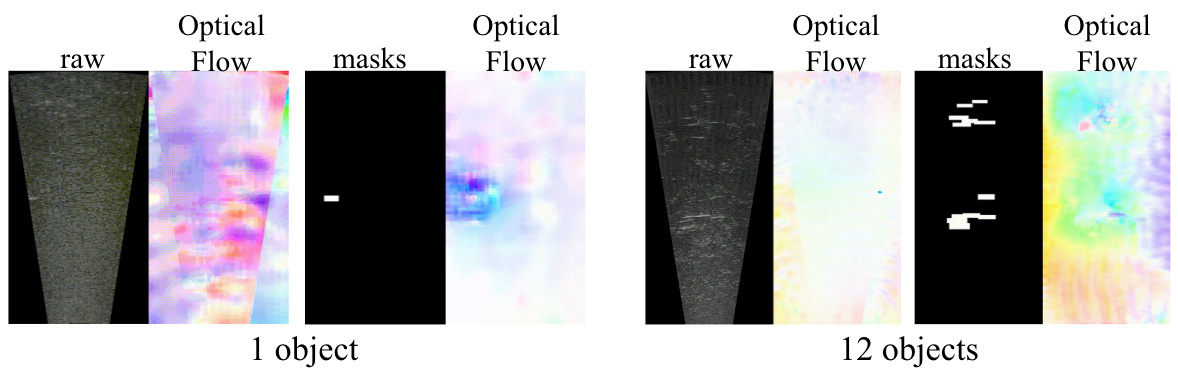}
    \caption{\textbf{RAFT \cite{teed2020RAFT} on CFC22 imagery and bounding box masks.} On the left, we can see the optical flow signal does not find the fish. When looking at the motion from the bounding-box mask of the fish (making the background movement stationary), the optical flow signal area is far greater than the actual area of the fish. On the right are frames (with fish and corresponding bounding-box masks), when there are 12 fish in the frame at once. Again, optical flow's signal is weak with the fish movement compared to the background. With the mask movement, optical flow signals cluster in groups of masked fish, but individuals are difficult to distinguish.}
    \label{fig:RAFT_on_cfc}

\end{figure}

\end{document}